\documentclass{article}

\usepackage[preprint, nonatbib]{neurips_2024}
\pdfoutput=1
\usepackage{enumitem}

\usepackage[utf8]{inputenc} 
\usepackage[T1]{fontenc}    
\usepackage{url}            
\usepackage{booktabs}       
\usepackage{amsfonts}       
\usepackage{nicefrac}       
\usepackage{microtype}      
\usepackage{xcolor}         

\usepackage{bbding}         
\usepackage{graphicx}
\usepackage{tabularx}       
\usepackage{comment}
\usepackage{soul}
\usepackage{amsmath}
\usepackage{amssymb}
\usepackage{float}
\usepackage[symbol]{footmisc}
\usepackage{multirow}
\usepackage{xcolor}
\usepackage[colorlinks=true,
            linkcolor=blue,
            urlcolor=red,
            citecolor=blue,
            anchorcolor=blue
            ]{hyperref}
\usepackage{blindtext}

\usepackage[capitalize]{cleveref}
\crefname{section}{Sec.}{Secs.}
\Crefname{section}{Section}{Sections}
\Crefname{table}{Table}{Tables}
\crefname{table}{Tab.}{Tabs.}

\renewcommand{\special}[1]{\textcolor[RGB]{231,138,195}{#1}}
\renewcommand{\paragraph}[1]{\vspace{0.1em}\noindent\textbf{#1}}

\title{
    \begin{minipage}{0.1\textwidth}
        \centering
        \includegraphics[width=\textwidth]{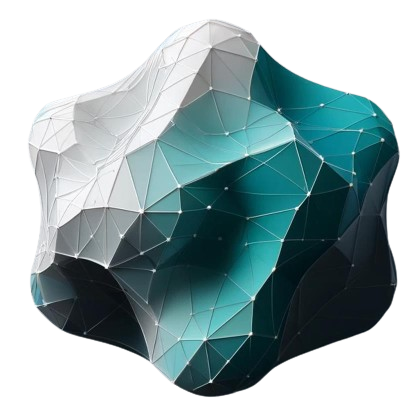}
    \end{minipage}%
    \hspace{-20pt}
    \begin{minipage}{0.8\textwidth}
        \centering
        MeshXL: Neural Coordinate Field for \\ 
        Generative 3D Foundation Models
    \end{minipage}
}

\author{
    \small{
        Sijin Chen$^{1,2,}$\thanks{Research done when Sijin Chen was a Research Intern at Tencent PCG.} , 
        Xin Chen$^{2,\dagger}$,
        Anqi Pang$^{2}$,
        Xianfang Zeng$^{2}$,
        Wei Cheng$^{2}$, 
        Yijun Fu$^{2}$,
        Fukun Yin$^{1,2}$
    } \\
    \textbf{
        \small{
            Yanru Wang$^{2}$,
            Zhibin Wang$^{2}$,
            Chi Zhang$^{2}$,
            Jingyi Yu$^{3}$,
            Gang Yu$^{2}$,
            Bin Fu$^{2}$,
            Tao Chen$^{1,\ddagger}$
        }
    }
    \\
    \tt \small \textbf{
        \href{https://github.com/OpenMeshLab/MeshXL}{https://github.com/OpenMeshLab/MeshXL}
    }
    \\
    {\normalsize $^{1}$Fudan University} \quad
    {\normalsize $^{2}$Tencent PCG} \quad
    {\normalsize $^{3}$ShanghaiTech University}
    \\
    {\small $^{\dagger}$ project lead} \quad {\small $^{\ddagger}$ corresponding author}
}

\begin{document}

\maketitle

\vspace{-5mm}
\begin{figure}[h]
    \vspace{-10pt}
    \centering
    \makebox[\textwidth][c]{
        \includegraphics[width=1.05\textwidth]{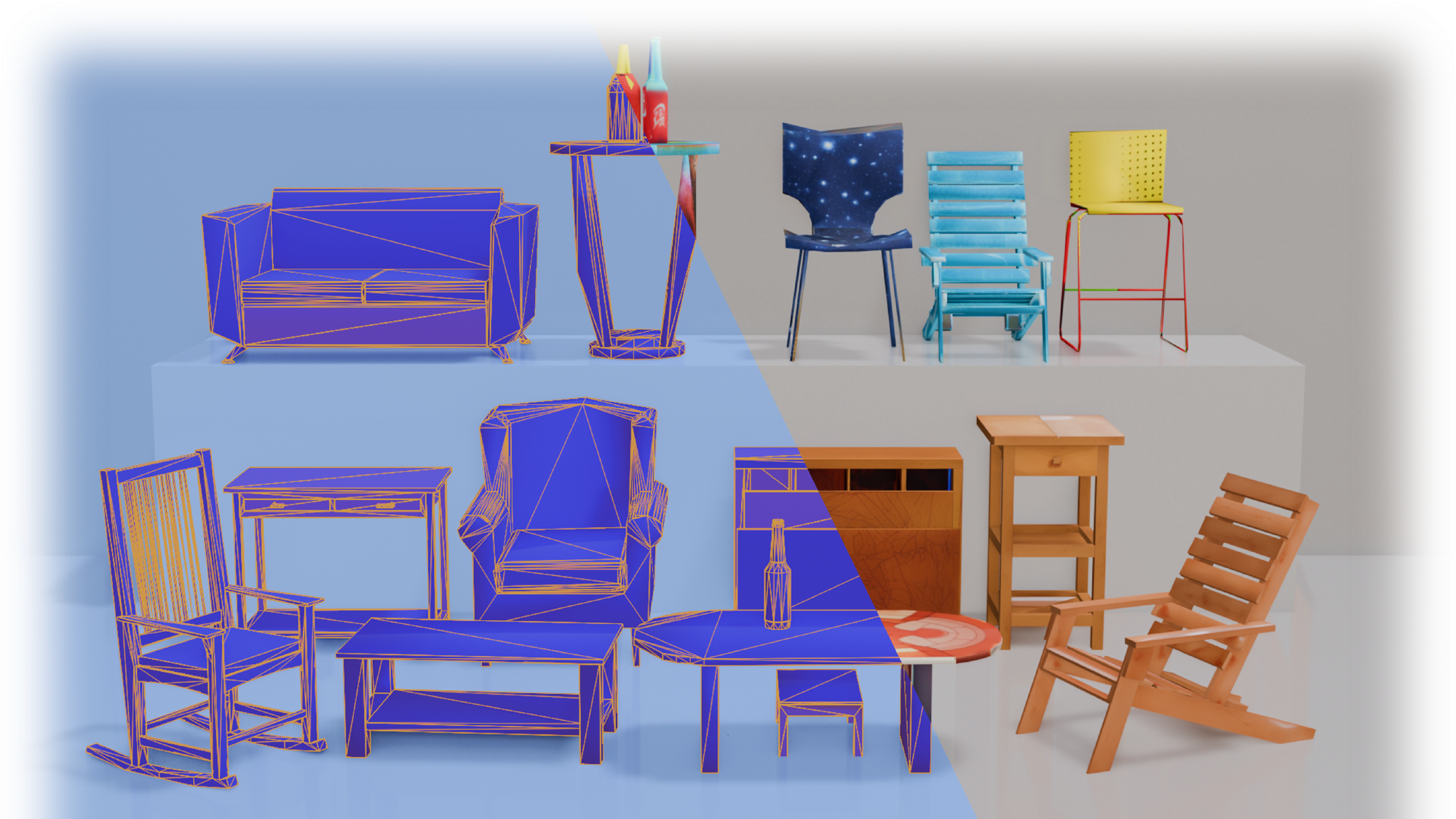}
    }
    \caption{
        \textbf{MeshXL} can auto-regressively generate high-quality 3D meshes.
        We validate that \textbf{Neur}al \textbf{C}oordinate \textbf{F}ield (NeurCF), an explicit coordinate representation with implicit neural embeddings, is a simple-yet-effective sequence representation for large-scale mesh modelling.
    }
    \label{fig:teaser}
\end{figure}

\begin{abstract}

    The polygon mesh representation of 3D data exhibits great flexibility, fast rendering speed, and storage efficiency, which is widely preferred in various applications.
    However, given its unstructured graph representation, the direct generation of high-fidelity 3D meshes is challenging.
    Fortunately, with a pre-defined ordering strategy, 3D meshes can be represented as sequences, and the generation process can be seamlessly treated as an auto-regressive problem.
    In this paper, we validate the \textbf{Neur}al \textbf{C}oordinate \textbf{F}ield (NeurCF), an explicit coordinate representation with implicit neural embeddings, is a simple-yet-effective representation for large-scale sequential mesh modeling.
    After that, we present MeshXL, a family of generative pre-trained auto-regressive models, which addresses the process of 3D mesh generation with modern large language model approaches.
    Extensive experiments show that MeshXL is able to generate high-quality 3D meshes, and can also serve as foundation models for various down-stream applications.
    
\end{abstract}

\section{Introduction}

The generation of high-quality 3D assets~\cite{ren2023xcube,wang2024themestation,hong20243dtopia} is essential for various applications in video games, virtual reality, and robotics.
Among existing 3D representations~\cite{mildenhall2020nerf,kerbl20233dgs,qi2017pointnet,ren2023xcube}, the 3D mesh represents the 3D data with graphs, which has the flexibility and accuracy for sharp edges as well as both flat and curved surfaces.
However, the direct generation of high-quality 3D meshes is challenging, given 1) the unstructured graph representation and 2) the demand for accurate spatial locations and connectivity estimation within vertices.

%
To generate 3D meshes, many works adopt an indirect way by first producing data in other 3D representations, including point clouds~\cite{zhou2021pvd,luo2021diffusion,nichol2022point-e}, SDF~\cite{yin2023shapegpt,zhao2024michelangelo}, and multi-view images~\cite{liu2023zero,xu2024instantmesh,hong2023lrm}.
After that, re-meshing methods~\cite{kazhdan2006poisson} are required for post-processing the generated geometries.
There are also attempts towards the direct generation of 3D polynomial meshes.
PolyGen~\cite{nash2020polygen} adopts two separate decoder-only transformers for vertices generation and connectivity prediction.
MeshGPT~\cite{siddiqui2023meshgpt} builds a mesh VQVAE to reconstruct the tokens generated by a GPT model~\cite{radford2019gpt2} into 3D meshes.
Meanwhile, PolyDiff~\cite{alliegro2023polydiff} directly adopts discrete denoising diffusion~\cite{austin2021discrete-diffusion} on the discretized mesh coordinates.

%
Though these methods have achieved initial success in 3D assets generation, they suffer from certain limitations.
To preserve high-frequency information, the point cloud and voxel representations will make dense samplings on the object surfaces, which inevitably lead to great redundancy when representing flat surfaces.
The reconstruction-based methods~\cite{xu2024instantmesh,hong2023lrm,tang2024lgm}, however, rely heavily on the quality of the multi-vew generation pipeline~\cite{liu2023zero}.
Additionally, the VQVAE-based 3D generation methods~\cite{yin2023shapegpt,siddiqui2023meshgpt} will inevitably result in cumulative errors when reconstructing the generated tokens into 3D structures.
%

To tackle the above challenges and explore the potential of scaling up 3D generative pre-training, we introduce a simple-yet-effective way of 3D mesh representation, the \textbf{Neur}al \textbf{C}oordinate \textbf{F}ield (NeurCF).
NeurCF represents the explicit 3D coordinates with implicit neural embeddings.
We show that with a pre-defined ordering strategy, the generation of 3D meshes can be formulated as an auto-regressive problem.
After that, we present MeshXL, a family of generative pre-trained transformers~\cite{zhang2022opt,radford2019gpt2}, for the direct generation of high-fidelity 3D meshes.
Without resorting to intermediate 3D representations, NeurCF facilitates an end-to-end learning pipeline for the direct pre-training on large-scale 3D mesh data.

By organizing high-quality 3D assets from ShapeNet~\cite{chang2015shapenet}, 3D-FUTURE~\cite{fu20213d-future}, Objaverse~\cite{deitke2023objaverse}, and Objaverse-XL~\cite{deitke2024objaverse-xl}, we achieve a collection of over 2.5 million 3D meshes to support large-scale generative pre-training.
Extensive experiments demonstrate that the NeurCF representation facilitates MeshXL to generate higher-quality 3D meshes with an increased number of parameters and large-scale pre-training data.
By training on the collection of large-scale 3D mesh data, MeshXL can achieve better performance with larger numbers of parameters (\cref{fig:scale-law} and \cref{tab:scale-study}),
and surpass prior arts on multiple categories task of the ShapeNet dataset~\cite{chang2015shapenet} (\cref{tab:full-comparison}).

In summary, our contributions can be summarized as follows:

\begin{itemize}[leftmargin=*]
    \item We validate that Neural Coordinate Field is a simple-and-effective representation of 3D mesh, which is also friendly to large-scale auto-regressive pre-training.

    \item We present a family of MeshXLs that can be treated as strong base models for image-conditioned or text-conditioned 3D mesh generation tasks.

    \item We show that MeshXL surpasses state-of-the-art 3D mesh generation methods, and can produce delicate 3D meshes compatible with existing texturing methods.
    
\end{itemize}
\section{Related Work}

First, we present a concise review of existing 3D representations. 
Subsequently, we discuss related works on 3D generation and recent efforts in developing 3D foundation models.

\paragraph{3D Representations.}
Researchers have long sought for accurate and efficient methods to represent 3D data. 
\textbf{Point Cloud}~\cite{nichol2022point-e,qi2017pointnet,qi2017pointnet++,yu2023pushing} captures the spatial positions of discrete points in the Euclidean space, which is preferred by various 3D sensors~\cite{dai2017scannet,yeshwanth2023scannet++,song2015sun,armeni20163d,behley2019semantickitti}.
\textbf{Mesh}~\cite{nash2020polygen,alliegro2023polydiff,siddiqui2023meshgpt,chen2020bsp} represents the 3D structure with graphs. By connecting the vertices with edges, mesh can also be interpreted into a set of polygons in the 3D space.
Similar to point clouds, \textbf{3D Gaussians}~\cite{kerbl20233dgs,tang2023dreamgaussian} also record the discrete Euclidean distribution in 3D space. However, each point is represented by a 3D Gaussian distribution function parameterized by its covariance matrix, color, and opacity. Given their fast convergence and rendering speed, 3D gaussians are often utilized for 3D reconstruction.
\textbf{Neural Radiance Field} (NeRF)~\cite{mildenhall2020nerf,barron2021mip-nerf} constructs a learnable volumetric function $f$ using neural networks trained on multi-view images. Due to its derivability and flexibility, NeRF is also favored for 3D generative models~\cite{liu2023zero,zhu2023hifa,wang2024prolificdreamer,poole2022dreamfusion}. Additionally, there are other 3D representations such as multi-view images~\cite{wang2023embodiedscan,yu2023mvimgnet,zhu2023pointclipv2}, voxel fields~\cite{ren2023xcube,choy2019Minkowski,liu2024one-2-3-45}, and signed distance fields~\cite{zhao2024michelangelo}, among others~\cite{shue20233d-triplane-diff,yin2023shapegpt,shen2021dmtet}. In this paper, we consider the \textbf{Neural Coordinate Field} (NeurCF), an explicit spatial representation with implicit neural embeddings, and investigate its potential for scalable 3D asset generation.

\paragraph{3D Generation.}
With the exploration of various 3D representations and the collection of large-scale 3D datasets~\cite{deitke2023objaverse,chang2015shapenet,deitke2024objaverse-xl}, researchers have also put much effort exploring the generation of high-fidelity 3D assets~\cite{li2024advances,li2023generative}.
The \textbf{G}enerative \textbf{A}dversarial \textbf{N}etwork (GAN)~\cite{goodfellow2020generative,wu2016-3dgan,achlioptas2018learning,ibing20213d} produces synthetic 3D data with a generator $\mathcal{G}$, and train a discriminator network $\mathcal{D}$ to distinguish the generated and real data.
Additionally, the potential of \textbf{diffusion} models~\cite{nichol2022point-e,ho2020ddpm,rombach2022high} in the direct generation of 3D data is also widely explored~\cite{zhou2021pvd,alliegro2023polydiff,nichol2022point-e,lyu2023controllable,liu2023meshdiffusion}.
The key idea behind diffusion is to transform the desired data distribution into a simpler distribution (\textit{e.g.} gaussian) and learn a desnoising model for the reverse process.
Besides, researchers have also explored the potential of diffusion models in generating \textbf{multi-view} images~\cite{liu2023zero,deitke2024objaverse-xl,xu2024instantmesh,liu2023one-2-3-45++}, and reconstruct them into 3D structures.
In this paper, we mainly explore the \textbf{auto-regressive} methods for 3D generation.
AutoSDF~\cite{mittal2022autosdf} and MeshGPT~\cite{siddiqui2023meshgpt} learn to generate discrete tokens and reconstruct them into 3D representations with a VQVAE model~\cite{van2017vqvae}.
PolyGen~\cite{nash2020polygen} adopts two decoder-only transformers that predict the location and connectivity of vertices, sequentially.
In this paper, we explore the potential of an explicit sequential modelling method for 3D meshes, and present a family of generative pre-trained transformers, MeshXL, for high-fidelity 3D mesh generation.

\paragraph{3D Foundation Models.} 
The collection of large-scale high-quality 3D data~\cite{deitke2023objaverse,deitke2024objaverse-xl,chang2015shapenet,wu2015modelnet,uy2019scanobjectnn,fu20213d-front,fu20213d-future} builds up the foundation for various 3D-related tasks~\cite{xu2023pointllm,guo2023pointbind_pointllm,chen2023ll3da,li2023m3dbench}.
To explore the scaling effects in 3D learning, researchers have made great endeavors in building 3D foundation models for 3D understanding~\cite{zhou2023uni3d,liu2024openshape,zhu2023ponderv2,xue2023ulip,xue2023ulip2,zhang2022pointclip,zhu2023pointclipv2}, 
reconstruction~\cite{hong2023lrm,wei2024meshlrm,tang2024lgm,liu2023zero,deitke2024objaverse-xl,xu2024grm,wang2023pf}, and 
generation~\cite{ren2023xcube,hong20243dtopia,siddiqui2023meshgpt,cao2024difftf++}.
With the introduction of large-scale 3D data in both variety and granularity~\cite{jia2024sceneverse,li2023m3dbench,deitke2024objaverse-xl}, existing 3D foundation models are capable of generalizing to 
unseen concepts~\cite{zhu2023pointclipv2,xue2023ulip2,liu2024openshape}, 
generating high-fidelity 3D assets~\cite{yin2023shapegpt,jiang2024motiongpt,siddiqui2023meshgpt}, 
responding to complex instructions~\cite{hong20233d-llm,chen2023ll3da,huang2023embodied,li2023m3dbench}, 
and generating actions that interacts with the 3D environments~\cite{driess2023palm,wu2023unleashing,zhen20243d-vla}.
In this paper, we present a fully end-to-end 3D mesh generation pipeline, explore the scaling effect for large-scale pre-training, and test whether our method can serve as a well-trained foundation model for various down-stream tasks.
\section{Data}
\label{supp-sec:data-source}

\paragraph{Data Sources.} 
We provide details on the 3D data collections we use to train and evaluate our models.
The whole data collection is built upon four widely-acknowledged 3D mesh datasets, \textit{i.e.} ShapeNet V2~\cite{chang2015shapenet}, 3D-FUTURE~\cite{fu20213d-future}, Objaverse~\cite{deitke2023objaverse}, and Objaverse-XL~\cite{deitke2024objaverse-xl}.

\begin{itemize}[leftmargin=*]
    \item  \textbf{ShapeNet V2~\cite{chang2015shapenet}} collects about 51k 3D CAD models for 55 categories.
    We split the data in 9:1 for training and validation by each category.

    \item \textbf{3D-FUTURE~\cite{fu20213d-future}} present about 10k high-quality 3D mesh data for indoor furniture.
    However, because of the delicate design, the objects contain many faces.
    Therefore, only a small proportion of the data can be used to train our MeshXL models.

    \item \textbf{Objaverse~\cite{deitke2023objaverse}} is a large 3D data collection with more than 800k 3D objects for about 21k categories collected from Sketchfab.
    We split the data in 99:1 for training and validation, respectively.

    \item \textbf{Objaverse-XL~\cite{deitke2024objaverse-xl}} further expand Objaverse~\cite{deitke2023objaverse} into a dataset with more than 10M 3D objects with additional data collected from GitHub, Polycam, Thingiverse, and Smithsonian.
    We split the Github and Thingiverse part of the Objaverse-XL dataset into 99:1 for training and validation, respectively.

\end{itemize}

\paragraph{Data collection and filtering.}
To organize existing datasets, we build up a filtering and pre-processing pipeline to ensure that the meshes met our demand.
We first collect meshes with fewer than 800 faces, and ensure that they have corresponding UV maps for rendering.
After that, we render the 3D meshes, and discard those are not center-aligned or occupying less than 10\% of the frame.
For those 3D meshes with more than 800 but less than 20,000 faces, we use planar decimation whether their meshes can be simplified.
Finally, we achieve approximately 2.5 million pieces of data remained.

\paragraph{Planar Decimation Pipeline.}
To ensure the quality of the decimated 3D meshes, we make sure either a lower Hausdorff distance $\delta_{\text{hausdorff}}$~\cite{siddiqui2023meshgpt} or a similar rendered views~\cite{chen2023robust}.

\paragraph{Collecting mesh-text pairs.}
We first render each 3D mesh with 12 different views, and concatenate them into one single image.
Then, we annotate both the front view image and the fused multi-view image using CogVLM~\cite{wang2023cogvlm}. 
After that, we adopt the Mistral-7B-Instruct model~\cite{jiang2023mistral} with few-shot in-context examples to extract information on category and geometry from the CogVLM annotations.
We tag each 3D mesh with the resulting categories and 3 to 5 geometry descriptors.

\paragraph{Collecting mesh-image pairs.} 
To produce diverse image conditions for 3D mesh generation, we first generate images with multi-view image and depth rendering.
After that, we use the sentences produced by CogVLM~\cite{wang2023cogvlm} as the prompt, and use a find-tuned Stable Diffusion model~\cite{rombach2022latent-diffusion} to augment the rendered images for diverse textures and backgrounds.
To ensure the quality of the generated images, we also adopt a manually cleansing procedure.

\paragraph{Data Statistics.} 
We present the data statistics of our large-scale 3D mesh collection in \cref{tab:data}.
After organizing and combing 3D assets from ShapeNet~\cite{chang2015shapenet}, 3D-FUTURE~\cite{fu20213d-future}, Objaverse~\cite{deitke2023objaverse}, and Objaverse-XL~\cite{deitke2024objaverse-xl}, we could achieve a total of 2.5 million 3D meshes.

\begin{table}[htbp]
\centering
    \caption{
        \textbf{Statistics for the Training Data and Validation Data.}
        After combining four data sources, our proposed MeshXL models are trained on approximately 2.5 million 3D meshes.
    }
    \label{tab:data}
    \resizebox{0.7\linewidth}{!}{
    \begin{tabular}{lccccc}
    \toprule
    \multirow{2}{*}{Dataset}                       & \multicolumn{2}{c}{Pre-training} &  & \multicolumn{2}{c}{Text-to-3D}  \\ \cline{2-3} \cline{5-6}
                                                   & Train            & Val           &  & Train           & Val           \\ \hline
    ShapeNet~\cite{chang2015shapenet}              & 16,001           & 1,754         &  & 15,384           & 1,728        \\
    3D-Future~\cite{fu20213d-future}               & 1,603            & -             &  & -               & -             \\
    Objaverse~\cite{deitke2023objaverse}           & 85,282           & 854           &  & 83,501           & 820          \\
    Objaverse-XL~\cite{deitke2024objaverse-xl}     & 2,407,337        & 15,200        &  & 1,347,802         & 13,579      \\ \cline{1-1}
    \textbf{Total}                        & 2,510,223        & 17,808        &  & 1,446,678         & 16,127      \\
    \bottomrule
    \end{tabular}
}
\end{table}

\section{Neural Coordinate Field}
\label{sec:neural_coordinate_field}

\textbf{Neur}al \textbf{C}oordinate \textbf{F}ield (NeurCF) is an explicit representation with implicit neural embeddings.
To be specific, for a Euclidean 3D coordinate system, we can partition the vertices coordinates into an $N^3$ grid. 
Then, each discretized coordinate $p=(x, y, z)$ can be encoded with the coordinate embedding layer $\mathcal{E}$, where $\mathcal{F}(p) = \left(\mathcal{E}(x), \mathcal{E}(y), \mathcal{E}(z)\right)$.
Therefore, a $k$-sided polynomial face $f^{(i)}$ can be encoded with 
$\mathcal{E}_{\text{face}}(f^{(i)}) = (\mathcal{F}(p_{1}^{(i)}), \cdots, \mathcal{F}(p_{k}^{(i)}))$.
For simplicity, the learnable coordinate embeddings $\mathcal{E}$ are shared among axes.

\paragraph{Ordering.}
Due to the graph representation, the order of the mesh vertices and the order of the edges between them are permutation-invariant.
%
A pre-defined ordering strategy is essential to facilitate the sequence modelling in MeshXL.
We employ the same ordering strategy as PolyGen~\cite{nash2020polygen} and MeshGPT~\cite{siddiqui2023meshgpt}.
The mesh coordinates are first normalized into a unit cube based on the mesh's longest axis, and discretized into unsigned integers.
Within each face, the vertices are cyclically permuted based their coordinates ($z$-$y$-$x$ order, from lower to higher), which helps to preserve the direction of normal vectors.
Then, we order these faces based on the permuted coordinates (lower to high).
To this end, an $n$-faced 3D $k$-sided polynomial mesh can be represented as $\mathcal{M} \in \mathbb{Z}^{n \times k \times 3}$, and we can encode $\mathcal{M}$ with $\mathcal{E}_{\text{mesh}} = (\mathcal{E}_{\text{face}}(f^{(1)}), \cdots, \mathcal{E}_{\text{face}}(f^{(n)}))$.

\begin{figure}[htbp]
    \centering
    \includegraphics[width=\linewidth]{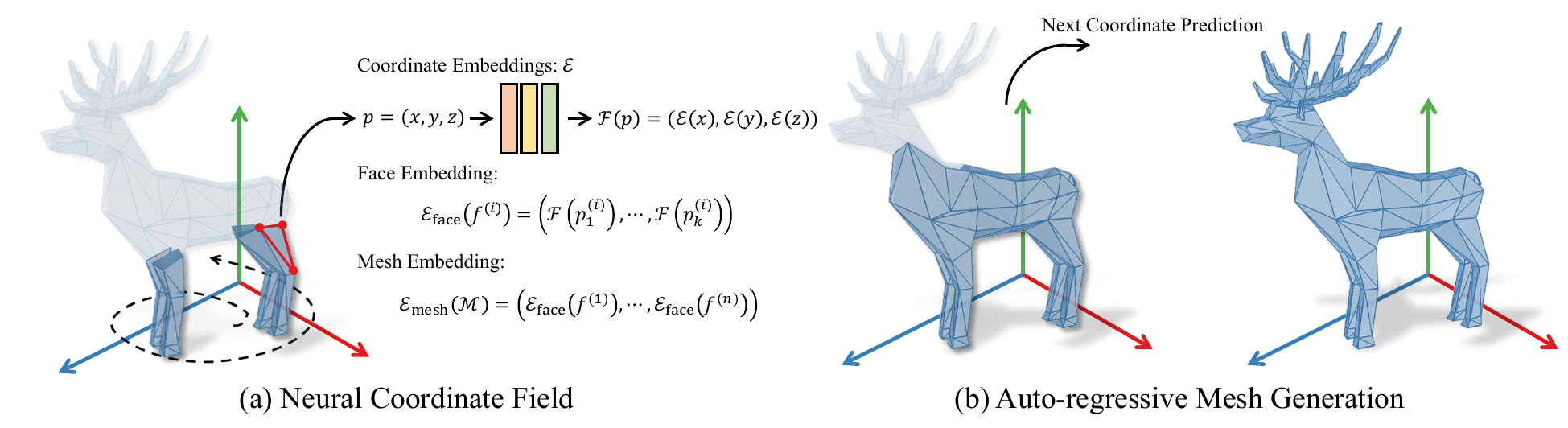}
    \caption{
        \textbf{Mesh Representation.}
        We present the \textbf{Neur}al \textbf{C}oordinate \textbf{F}ield (NeurCF) to encode the discretized coordinates in the Euclidean space.
        Benefiting from NeurCF and a pre-defined ordering strategy, our proposed MeshXL can directly generate the unstructured 3D mesh auto-regressively.
    }
    \label{fig:method}
\end{figure}

\paragraph{A Sequential Mesh Representation.} 
One direct way to represent the 3D meshes is to directly reshape $\mathcal{M}$ into a vector with $(n \cdot k \cdot 3)$ tokens.
As a special case, an $n$-faced triangular mesh can be represented by a vector with $9n$ tokens.
Meanwhile, our representation can also be expanded to hybrid polynomial mesh representations with the proper introduction of separate tokens.
For example, we can generate triangles within ``\special{<tri> $\cdots$ </tri>}'' and quadrilaterals within ``\special{<quad> $\cdots$ </quad>}''.
To identify the start and end of a mesh sequence, we add a \special{<bos>} (``begin-of-sequence'') token before the mesh sequence and an \special{<eos>} (``end-of-sequence'') token after.

\paragraph{Comparisons.}
Compared to other forms of 3D representations, NeurCF is a direct representation for 3D meshes.
Since we represent each coordinate with learnable embeddings, NeurCF is an end-to-end trainable representation for unstructured 3D meshes.
Additionally, NeurCF is storage efficient comparing to voxel fields ($O(N^3)$) and point clouds, since it can naturally model the flat surfaces with graph structures.

\section{Method}

We first present the architecture and training objective for MeshXL models.
Then, we show that MeshXL models can take an additional modality as the condition for controllable 3D assets generation.
After this, we investigate the effects of scaling.

\paragraph{Architecture.}
In \cref{sec:neural_coordinate_field}, we present a simple-yet-effective way to represent a 3D mesh into a sequence.
Therefore, the learning of 3D mesh generation can be formulated into an auto-regressive problem, and can be seaminglessly addressed by modern \textbf{L}arge \textbf{L}anguage \textbf{M}odel (LLM) approaches.
In our paper, we adopt the decoder-only transformers using the OPT~\cite{zhang2022opt} codebase as our base models.
To adapt the pre-trained OPT models to our \textit{next-coordinate prediction} setting, we fine-tune the whole model with newly-initialized coordinate and position embeddings.

\paragraph{Generative Pre-Training.}
We use the standard next-token prediction loss to train our models.
Given the trainable weights $\theta$ and an $|s|$-length sequence $s$, the generation loss is calculated as:
\begin{equation}
    \mathcal{L}_{\text{MeshXL}}\left(\theta\right) = - \sum_{i=1}^{|s|} \log P\left(s_{\left[i\right]} \vert s_{\left[1, \cdots, i-1\right]}; \theta \right).
    \label{eq:ce_loss}
\end{equation}

For each mesh sequence, we add a \special{<bos>} token before the mesh tokens, and an \special{<eos>} token after the mesh tokens to identify the ending of a 3D mesh.
During inference, we adopt the top-$k$ and top-$p$ sampling strategy to produce diverse outputs.

\begin{figure}[htbp]
    \centering
    \includegraphics[width=\linewidth]{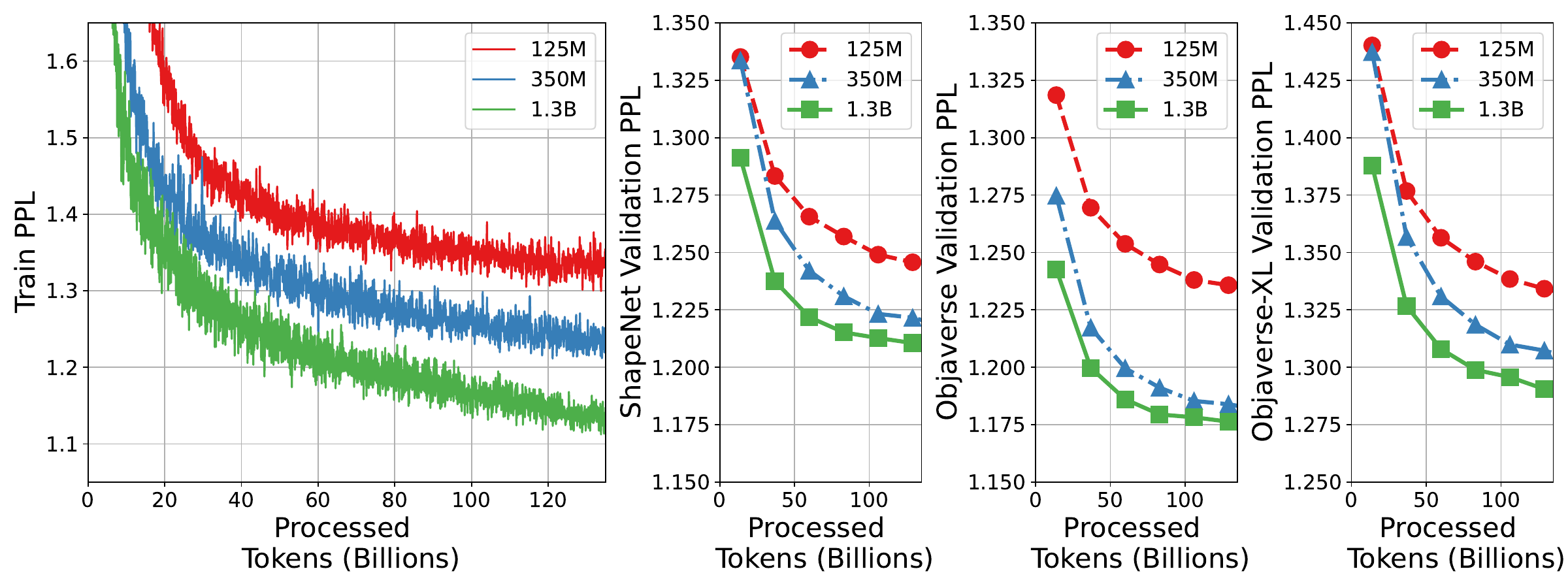}
    \caption{
        \textbf{Training and Validation Perplexity (PPL) for MeshXL Models.}
        We train all the models from scratch on 150 billion tokens.
        We observe that the performance grows with model sizes.
    }
    \label{fig:scale-law}
\end{figure}

\paragraph{$\mathcal{X}$-to-Mesh Generation.}
\label{subsec:condition_method}
Here we mainly consider generating 3D meshes from images and texts.
We adopt a pre-trained BERT~\cite{devlin2018bert} model for text feature encoding, and a pre-trained ViT~\cite{dosovitskiy2020vit} model for image feature encoding.
To align the additional text/image feature with the mesh coordinate field, we adopt the Q-Former architecture~\cite{li2023blip2} to compress the encoded feature into a fixed-length of 32 learnable tokens as the prefix of the MeshXL model.
The overall training objective of the conditional mesh generation is shown in \cref{eq:x-to-3D-ce-loss}:
\begin{equation}
    \mathcal{L}_{\mathcal{X}\text{-to-mesh}}\left(\theta\right) = - \sum_{i=1}^{|s|} \log P\left(s_{\left[i\right]} \vert s_{\left[1, \cdots, i-1\right]}; \mathcal{X}\right).
    \label{eq:x-to-3D-ce-loss}
\end{equation}
During inference, the model predicts the mesh tokens after the fixed-length prefix.

\paragraph{Scaling Up.}
We present MeshXL in various sizes, including 125M, 350M, and 1.3B.
The detailed hyperparameters for training different models can be found in \cref{tab:hyperparameters}.
To better analyze the scaling effects, we train all models from scratch on 150 billion tokens.
We provide both training curve and validation perplexity for different models in \cref{fig:scale-law}.
One can see that as the number of parameters grows, the model achieves a lower validation perplexity, indicating a higher probability to produce the validation data.

\section{Experiments}

We first briefly introduce the data, metrics, and implementation details in \cref{subsec:data_metrics_implementation}.
Then, we provide evaluations and comparisons on the generated meshes ($cf.$ \cref{subsec:evaluation}) and ablations ($cf.$ \cref{subsec:ablation}).
We also provide visualization results in \cref{subsec:viz}.

\subsection{Data, Metrics, and Implementation Details}
\label{subsec:data_metrics_implementation}

\paragraph{Data.} We pre-train the base model with 2.5 million 3D meshes collected from the combination of ShapeNet~\cite{chang2015shapenet}, 3D-FUTURE~\cite{fu20213d-future}, Objaverse~\cite{deitke2023objaverse}, and Objaverse-XL~\cite{deitke2024objaverse-xl}.
We use planar decimation on meshes with more than 800 faces following MeshGPT~\cite{siddiqui2023meshgpt} and RobustLowPoly~\cite{chen2023robust}.
More details on the data collection and processing pipeline can be found in the appendix.
For generative mesh pre-training, we randomly rotate these meshes with degrees from (0$^{\circ}$, 90$^\circ$, 180$^\circ$, 270$^{\circ}$), and adopt random scaling along each axis within range $[0.9, 1.1]$ for data augmentation.

\paragraph{Metrics.} 
We follow the standard evaluation protocols in MeshGPT~\cite{siddiqui2023meshgpt} and PolyDiff~\cite{alliegro2023polydiff} with the following metrics.
Coverage (COV) is sensitive to mode dropping and is used to quantify the diversity of the generated meshes.
However, COV does not assess the quality of the generated results. 
Minimum Matching Distance (MMD) calculates the average distance between the reference set and their closest neighbors in the generated set.
However, MMD is not sensitive to low-quality results. 
The 1-Nearest Neighbor Accuracy (1-NNA) directly quantifies the quality and diversity between the generation set and the reference set.
The optimal value of 1-NNA is 50\%. 
We adopt the Jensen-Shannon Divergence (JSD) score to directly evaluate 3D meshes. 
We use Chamfer Distance to measure the similarity between two samples.
We also adopt the Frechet Inception Distance (FID) and Kernel Inception Distance (KID) on the rendered images for feature-level evaluation.
The MMD, JSD, and KID scores are multiplied by $10^3$.

\paragraph{Implementation.} 
All experiments are conducted on a cluster consisting of 128 A100 GPUs. 
We train our models under bfloat16 and the ZeRO-2 strategy~\cite{rajbhandari2020zero} using the AdamW~\cite{loshchilov2017adamw} optimizer with a learning rate decaying from $10^{-4}$ to $10^{-6}$ and a weight decay of 0.1.
The detailed hyperparameters for different models can be found in \cref{tab:hyperparameters}.
To train our base models, we load the weights from the pre-trained OPT models~\cite{zhang2022opt} and initialize the word embeddings and positional embeddings from scratch.
Without further specification, we generate 3D meshes with the top-$k$ and top-$p$ sampling strategy with $k=50$ and $p=0.95$.

\begin{table}[ht]
\centering
    \caption{
        \textbf{Hyperparameters for different MeshXL Base Models.} 
        We present three MeshXL models with 125M, 350M, and 1.3B parameters, respectively.
    }
    \label{tab:hyperparameters}
    \resizebox{0.8\linewidth}{!}{
        \begin{tabular}{lccc}
        \toprule
        Hyperparameters            & MeshXL(125M)        & MeshXL(350M)        & MeshXL(1.3B)        \\ \hline
        \# Layers                  & 12                   & 24                   & 24                   \\
        \# Heads                   & 12                   & 16                   & 32                   \\
        $d_{\text{model}}$         & 768                  & 1,024                & 2,048                \\
        $d_{\text{FFN}}$           & 3,072                & 4,096                & 8,192                \\
        Optimizer                  & \multicolumn{3}{c}{AdamW($\beta_1$=0.9, $\beta_2$=0.999)}          \\
        Learning rate              & $1.0 \times 10^{-4}$ & $1.0 \times 10^{-4}$ & $1.0 \times 10^{-4}$ \\
        LR scheduler               & Cosine               & Cosine               & Cosine               \\
        Weight decay               & 0.1                  & 0.1                  & 0.1                  \\
        Gradient Clip              & 1.0                  & 1.0                  & 1.0                  \\
        Number of GPUs             & 8                    & 16                   & 32                   \\
        \# GPU hrs (A100)          & 1,944                & 6,000                & 23,232               \\
        \bottomrule
        \end{tabular}
    }
\end{table}

\subsection{Evaluations and Comparisons}
\label{subsec:evaluation}

We provide quantitative as well as qualitative comparisons on both unconditional and conditional 3D mesh generation on public benchmarks.

\paragraph{Unconditional Generation.}
We evaluate MeshXL as well as other baseline methods using the ShapeNet~\cite{chang2015shapenet} data in \cref{tab:full-comparison}.
We split the data by 9:1 for training and validation by each category.
For evaluation, we fine-tune our pre-trained base model and sample 1,000 meshes for each category.
Among the listed methods, we reproduce the MeshGPT~\cite{siddiqui2023meshgpt} with a GPT2-medium model (355M)~\cite{radford2019gpt2}.
With a similar number of parameters, Mesh-XL (350M) out-performs MeshGPT by a large margin, showing a higher COV score, a lower MMD score, and a closer 1-NNA score to 50\%.
This indicates that MeshXL can produce diverse and high-quality 3D meshes.

\begin{table}[htbp]
    \centering
    \caption{
        \textbf{Quantitative Comparisons with Prior Arts on ShapeNet~\cite{chang2015shapenet}.}
        We scale MMD, JSD, KID by $10^3$.
        MeshXL can produce diverse and high-quality 3D meshes.
    }
    \label{tab:full-comparison}
    \resizebox{0.8\linewidth}{!}{
    \begin{tabular}{clcccccc}
    \toprule
    Category               & Methods                              & COV$\uparrow$ & MMD$\downarrow$ & 1-NNA & JSD$\downarrow$ & FID$\downarrow$ & KID$\downarrow$\\ 
    \bottomrule
    \multirow{6}{*}{Chair} & PolyGen~\cite{nash2020polygen}       &  7.79         & 16.00          & 99.16            & 228.80           & 63.49           & 43.73          \\
                           & GET3D~\cite{gao2022get3d}            & 11.70         & 15.92          & 99.75            & 155.25           & 67.84           & 42.10          \\
                           & MeshGPT~\cite{siddiqui2023meshgpt}   &  42.00         &  4.75          & 69.50            & 55.16           & 39.52           & 8.97          \\
                           \cline{2-8}
                           & MeshXL (125M)                       & 50.80          & \textbf{3.11}  & 56.55            & 9.69            & 28.15           & 1.48 \\
                           & MeshXL (350M)                       & 50.80          & 3.17           & \textbf{55.80}   & 9.66            & 28.29           & \textbf{1.39}          \\
                           & MeshXL (1.3B)                       & \textbf{51.60} & 3.23           & \textbf{55.80}   & \textbf{9.48}   & \textbf{9.12}  & 1.84          \\
                           \bottomrule
    \multirow{6}{*}{Table} & PolyGen~\cite{nash2020polygen}       & 44.00         &  3.36          & 67.20            & 25.06            & 54.08           & 14.96          \\
                           & GET3D~\cite{gao2022get3d}            & 16.80         & 10.39          & 91.90            & 226.97           & 67.65           & 34.62          \\
                           & MeshGPT~\cite{siddiqui2023meshgpt}   & 34.30         &  6.51          & 75.05            & 92.88           & 53.75           & 7.75          \\
                           \cline{2-8}
                           & MeshXL (125M)                       & 51.21          & 2.96           & 57.96            & \textbf{12.82}   & 42.55           & \textbf{0.92}          \\
                           & MeshXL (350M)                       & 49.70          & 3.07           & \textbf{56.10}   & 13.64           & 43.43           & 1.27          \\
                           & MeshXL (1.3B)                       & \textbf{52.12} & \textbf{2.92}  & 56.80            & 14.93           & \textbf{22.29} & 2.03 \\
                           \bottomrule
    \multirow{5}{*}{Bench} & PolyGen~\cite{nash2020polygen}       & 31.15         &  4.01           & 83.23 & 55.25            & 70.53           & 12.1         \\
                           & MeshGPT~\cite{siddiqui2023meshgpt}   & 34.92         &  2.22          & 68.65            & 57.32           & 52.47           &  6.49          \\
                           \cline{2-8}
                           & MeshXL (125M)                       & 54.37          & 1.65            & \textbf{43.75} & 16.43           & \textbf{35.31}  & \textbf{0.82} \\
                           & MeshXL (350M)                       & 53.37          & 1.65            & 42.96          & \textbf{15.41}  & 36.35           & 0.96          \\
                           & MeshXL (1.3B)                       & \textbf{56.55} & \textbf{1.62}   & 39.78          & 15.51           & 35.50           & 1.60          \\
                           \bottomrule
    \multirow{5}{*}{Lamp}  & PolyGen~\cite{nash2020polygen}       & 35.04         &  7.87           & 75.49          & 96.57            & 65.15           & 12.78          \\
                           & MeshGPT~\cite{siddiqui2023meshgpt}   & 41.59         &  4.92          & 61.59            & 61.82           & 47.19           & 5.19          \\
                           \cline{2-8}
                           & MeshXL (125M)                       & \textbf{55.86} & 5.06           & 48.24           & 43.41           & 34.61           & \textbf{0.84}          \\
                           & MeshXL (350M)                       & 53.52          & \textbf{4.18}  & \textbf{49.41}   & \textbf{34.87}  & \textbf{25.94}  & 1.92          \\
                           & MeshXL (1.3B)                       & 51.95          & 4.89           & 47.27            & 41.89           & 31.66           & 0.99 \\
                           \bottomrule
    \end{tabular}
}
\end{table}

\paragraph{User Study.} 
To evaluate how well the generated 3D meshes align with human preference, we perform user studies on the chair category in \cref{tab:user_study} with several baseline methods~\cite{nash2020polygen,gao2022get3d}.
We recruit and instruct the participants to score each mesh from 0 to 5 based on its 
1) \textbf{quality}: the smoothness of object surfaces and completeness of the mesh, 
2) \textbf{artistic}: how much do you believe this object is designed and created by artists, 
and 3) \textbf{triangulation}: how well do the connectivity among vertices aligns with the models created by professional designing software~\cite{blender}.
For the above mentioned metrics, the higher score means better quality.
As a baseline evaluation, we also ask the participants to score the ground truth 3D geometries sampled from the ShapeNet data.
We have collected a total of 434 valid responses.
The results show that the 3D meshes created by MeshXL are consistently preferred by human in all dimensions.

\begin{table}[htbp]
\centering
    \caption{
        \textbf{User Study.}
        Compared to baseline methods, the meshes generated by MeshXL are better aligned with human preference in terms of both geometry and designs.
    }
    \label{tab:user_study}
    \resizebox{0.55\linewidth}{!}{
    \begin{tabular}{lccc}
        \toprule
        Methods                            & Quality$\uparrow$         & Artistic$\uparrow$        & Triangulation$\uparrow$   \\ \hline
        PolyGen~\cite{nash2020polygen}     & 2.53            & 2.72            & 3.15            \\
        GET3D~\cite{gao2022get3d}          & 3.15            & 2.46            & 3.15            \\
        MeshXL                             & \textbf{3.96}   & \textbf{3.45}   & \textbf{3.72}   \\
        \cline{1-1}
        Reals    & 4.08             & 3.33            & 3.75   \\
        \bottomrule
    \end{tabular}
    }
\end{table}

\begin{figure}[t]
    \centering
    \includegraphics[width=\linewidth]{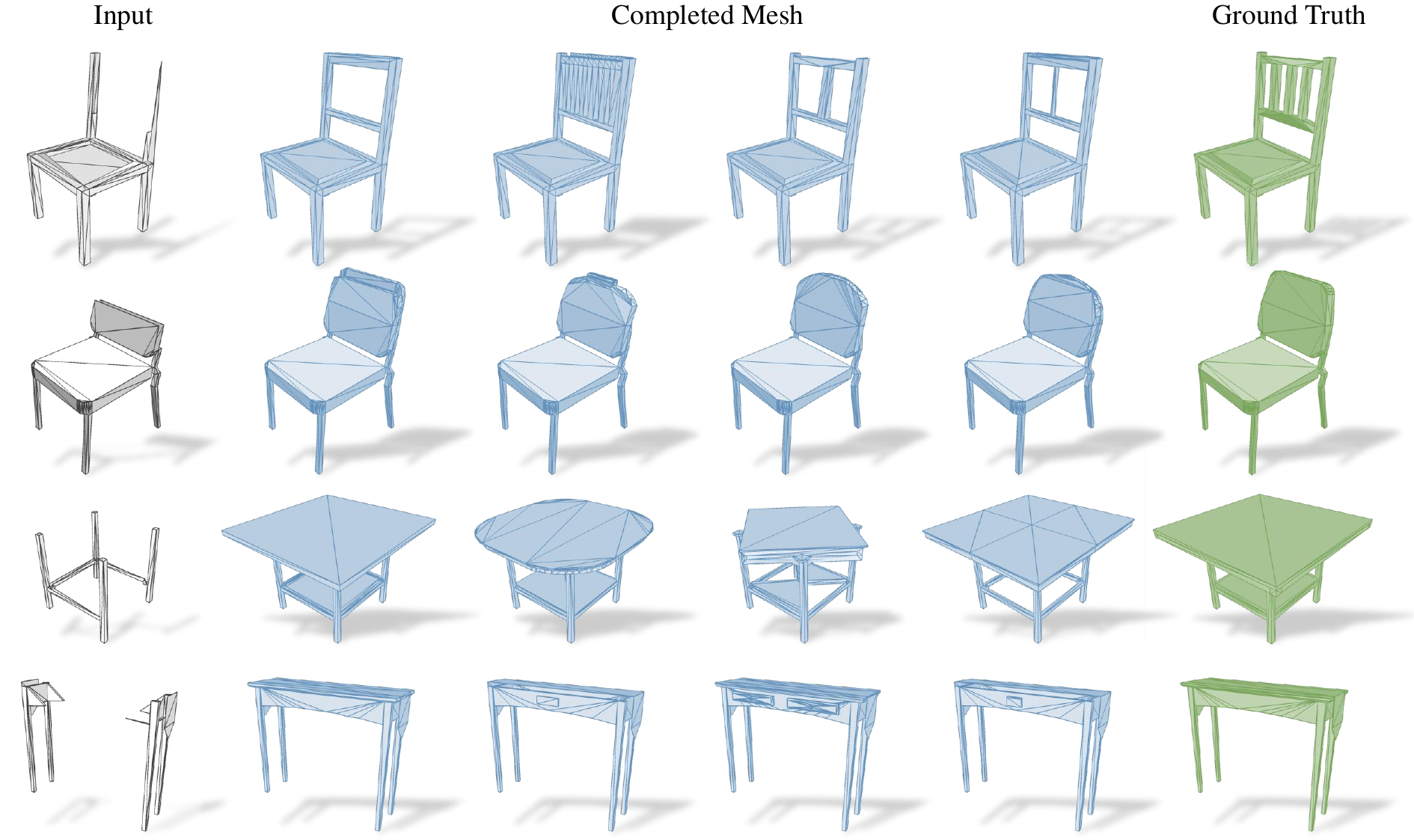}
    \caption{
        \textbf{Evaluation of Partial Mesh Completion.}
        Given some partial observation of the 3D mesh (white), MeshXL is able to produce diverse object completion results (blue).
    }
    \label{fig:completion_results}
\end{figure}

\subsection{Ablation Studies}
\label{subsec:ablation}

\paragraph{Necessity of Mesh VQVAE.} 
Comparing to MeshGPT~\cite{siddiqui2023meshgpt}, MeshXL is an end-to-end trainable model that produces 3D meshes with \textit{next-coordinate prediction}.
We show in \cref{tab:full-comparison} that, MeshXL out-performs MeshGPT with similar numbers of parameters.
Furthermore, MeshXL can save the effort training a mesh autoencoder~\cite{siddiqui2023meshgpt,van2017vqvae}, which further facilitates scaling up generative pre-training.

\paragraph{Shape Completion.}
To analysis whether our method is capable of producing diverse outputs, we ask MeshXL (1.3B) model to predict the whole object given some partial observations of the 3D mesh.
In practice, we use 50\% of the object mesh as input, and ask the model to predict the rest 50\% of the 3D mesh.
We illustrate completion examples on chairs and tables in \cref{fig:completion_results}.
One can see that Mesh-XL is able to produce diverse outputs given the partial observation of the 3D mesh.

\begin{figure}[t]
    \centering
    \includegraphics[width=1\linewidth]{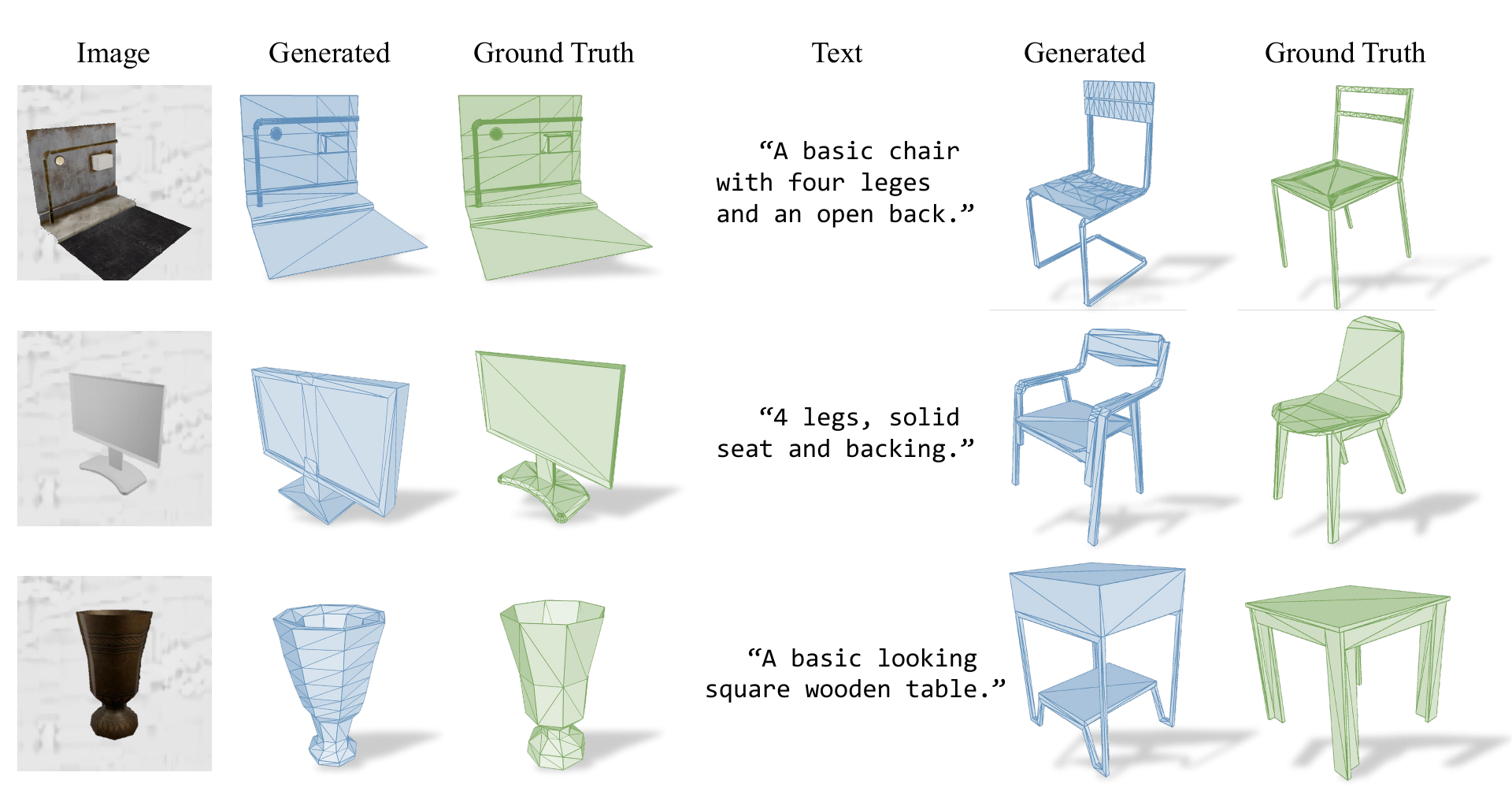}
    \vspace{-20pt}
    \caption{
        \textbf{Evaluation of $\mathcal{X}$-to-mesh generation.}
        We show that MeshXL can generate high-quality 3D meshes given the corresponding image or text as the additional inputs.
    }
    \label{fig:x-to-mesh}
\end{figure}

\paragraph{$\mathcal{X}$-to-Mesh Generation.}
We showcases several conditional generation results in \cref{fig:x-to-mesh}.
We show that MeshXL can generate high-quality 3D meshes given the corresponding image or text as the additional inputs.

\begin{figure}[t]
    \centering
    \includegraphics[width=\linewidth]{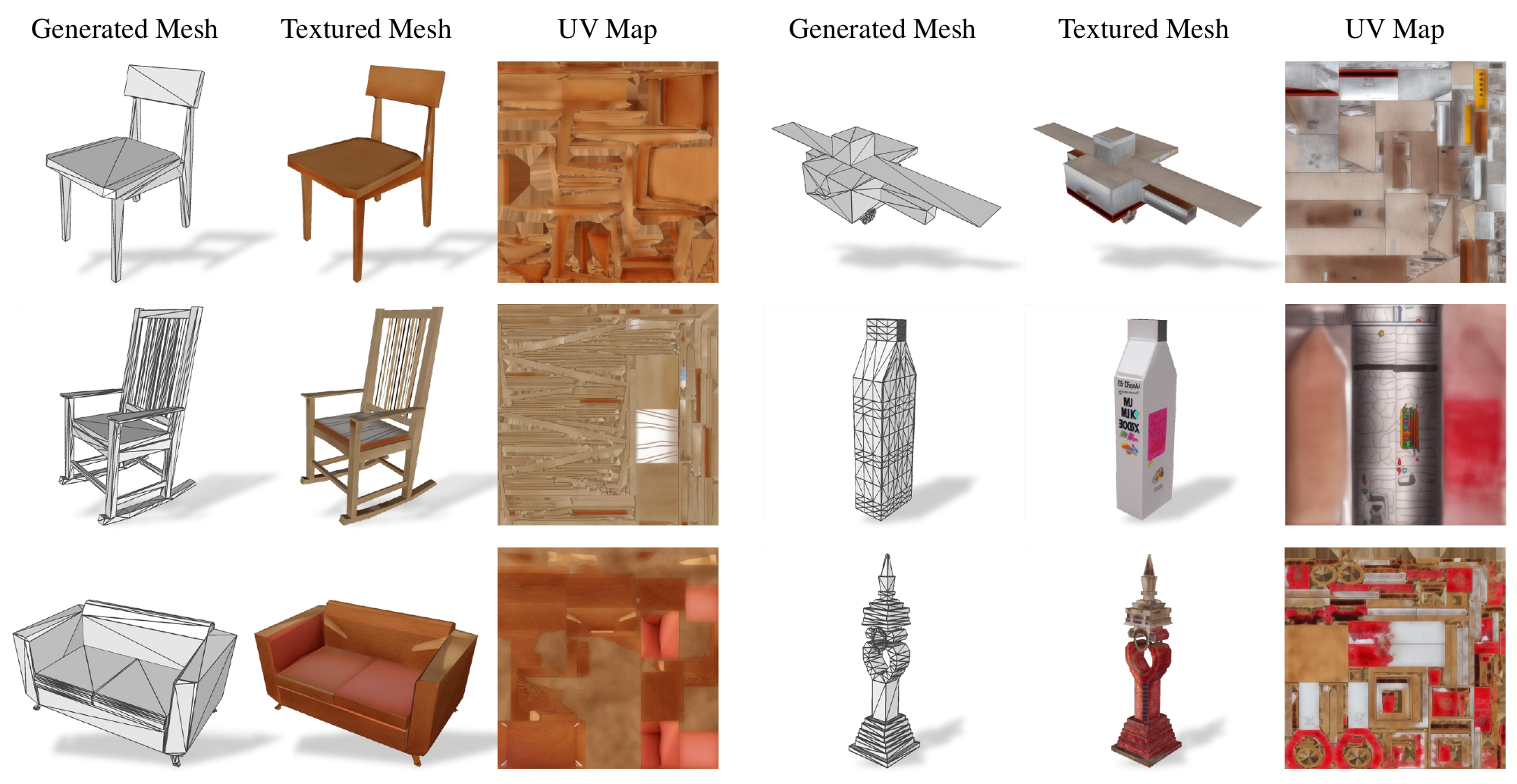}
    \vspace{-20pt}
    \caption{
        \textbf{Texture Generation for the Generated 3D Meshes.}
        We adopt Paint3D~\cite{zeng2023paint3d} to generate textures for 3D meshes produced by MeshXL.
    }
    \label{fig:texturing}
\end{figure}

\paragraph{Effectiveness of Model Sizes.} 
To analyze whether large-scale pre-training a larger model benefits 3D mesh generation, we evaluate MeshXL base models with different sizes on the Objaverse~\cite{deitke2023objaverse} dataset in \cref{tab:scale-study}.
We observe that as the model size grows, the generated samples exhibits a closer 1-NNA to 50\%, a larger COV, and smaller JSD score, which indicates an improving diversity and quality.

\begin{table*}[h]
    \centering
    \caption{
        \textbf{Effectiveness of Model Sizes on Objaverse.} 
        We observe that as the model size grows, the generated meshes exhibit a closer 1-NNA to 50\%, a larger COV and a smaller JSD, indicating better diversity and quality.
    }
    \label{tab:scale-study}
    \resizebox{0.7\linewidth}{!}{
    \begin{tabular}{lccccccccccccc}
        \toprule
        Method             & COV$\uparrow$  & MMD$\downarrow$ & 1-NNA          & JSD$\downarrow$ & FID$\downarrow$ & KID  $\downarrow$ \\ \hline
        MeshXL (125M)      & 39.76          & 5.21            & 67.34          & 26.03           & 17.32           & 4.48          \\
        MeshXL (350M)      & 40.79          & 5.20            & 65.68          & 23.71           & 15.14           & 3.33          \\
        MeshXL (1.3B)      & \textbf{42.86} & \textbf{4.16}   & \textbf{61.56} & \textbf{20.99}  & \textbf{12.49}           & \textbf{2.94}          \\
        \bottomrule
    \end{tabular}
}
\end{table*}

\paragraph{Texturing.}
We adopt Paint3D~\cite{zeng2023paint3d}, a coarse-to-fine texture generation pipeline, to generate textures for the 3D meshes produced by MeshXL in \cref{fig:texturing}.
We show that 3D meshes produced by MeshXL can easily fit the existing texturing methods to produce high-quality 3D assets.

\clearpage

\subsection{Visualizations}
\label{subsec:viz}

We provide qualitative comparisons on the meshes generated by our method as well as the meshes generated by other baseline models.

\begin{figure}[t]
    \centering
    \includegraphics[width=1\linewidth]{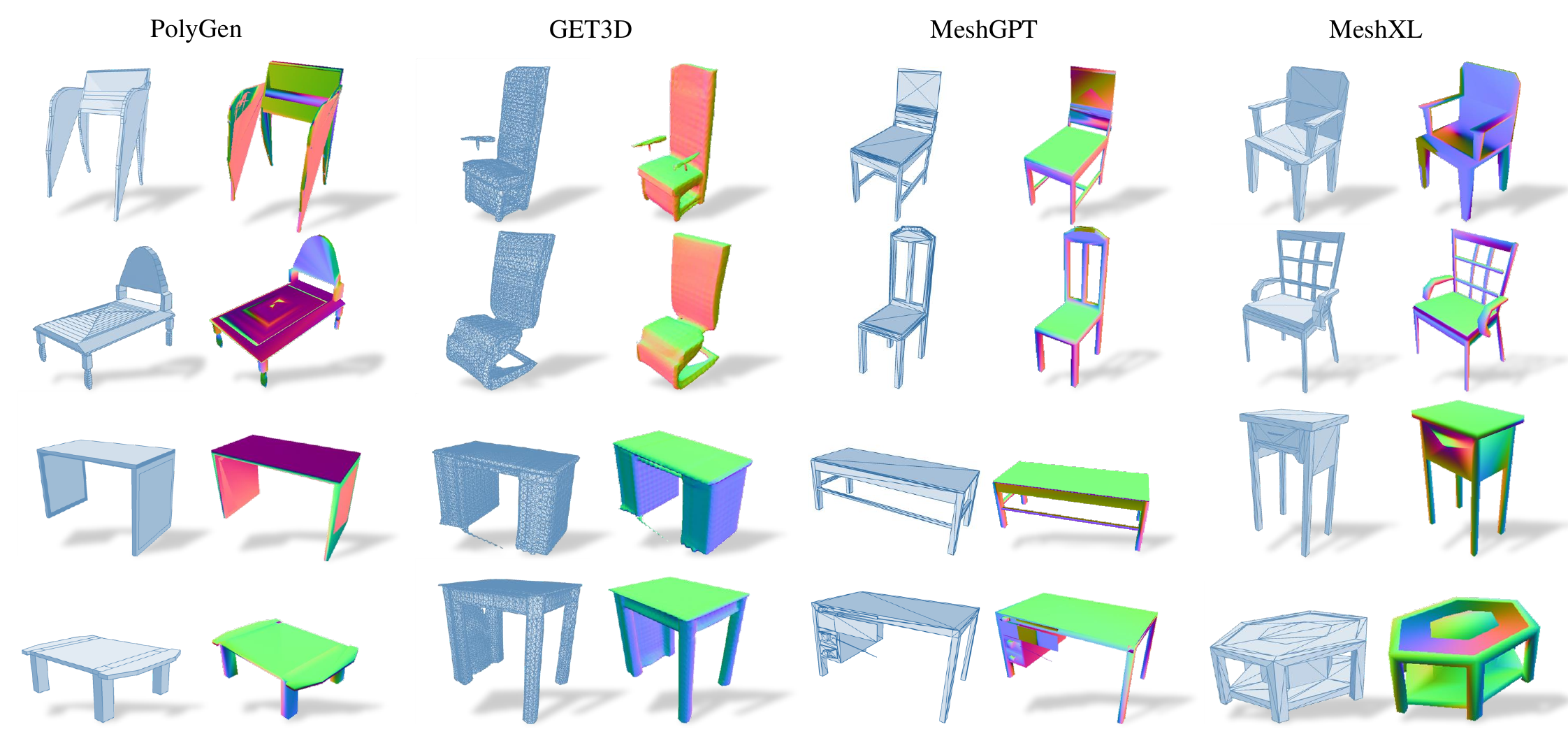}
    \caption{
        \textbf{Qualitative comparison on the generated meshes.}
        We present qualitative comparisons on the generated meshes as well as normal vectors.
        MeshXL is able to produce high-quality 3D meshes with both sharp edges and smooth surfaces.
    }
    \label{fig:qualitative-comparison}
\end{figure}

\paragraph{Qualitative Comparison.}
We provide category specified visualization results as well as their normal vectors on the generated meshes in \cref{fig:qualitative-comparison}.
With the ability to generate 3D meshes directly, MeshXL is able to produce high-quality 3D meshes with both sharp edges and smooth surfaces.

\begin{figure}[htbp]
    \centering
    \makebox[\textwidth][c]{
        \includegraphics[width=1.25\linewidth]{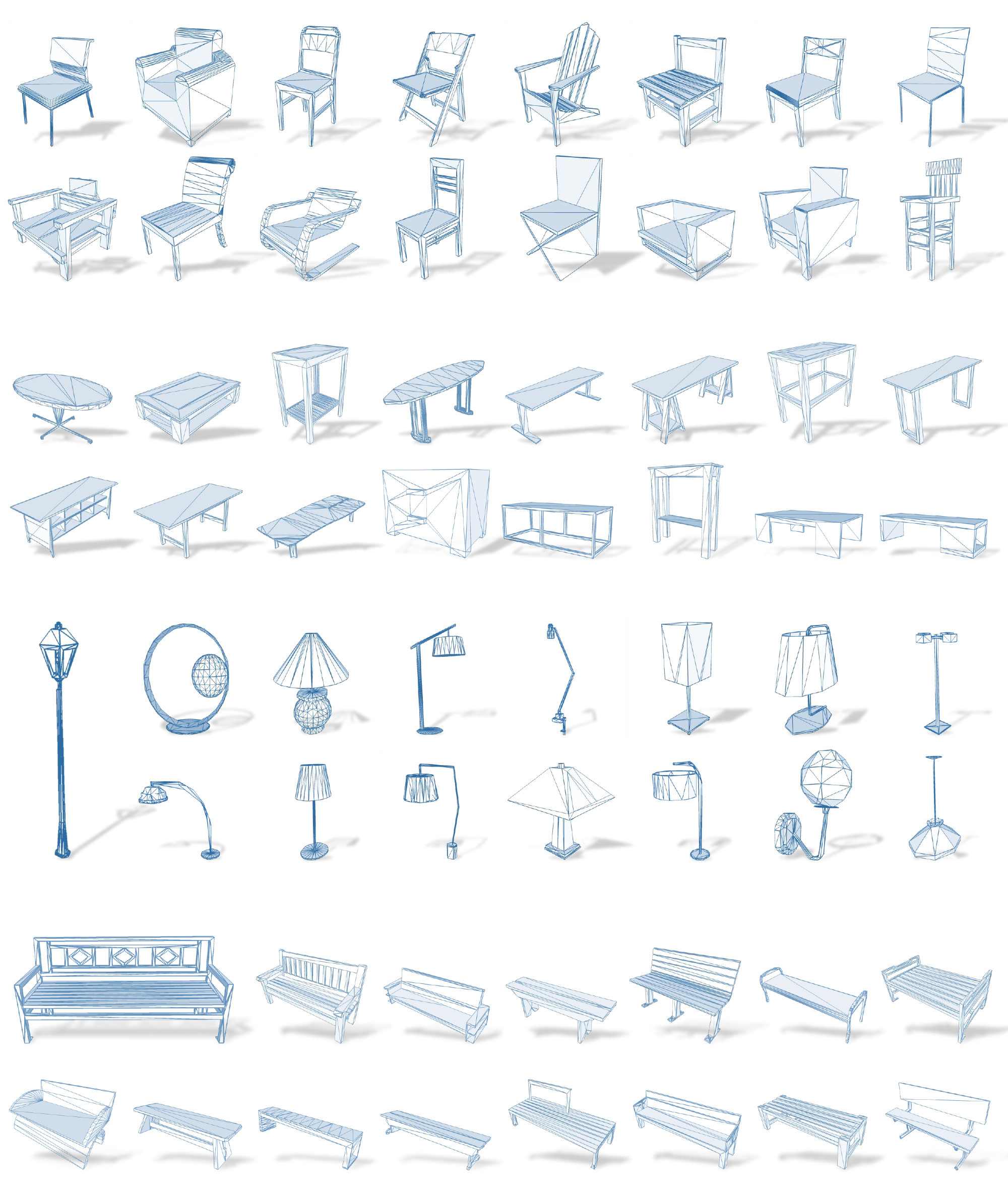}
    }
    \caption{
        \textbf{Gallery results.}
        Additional generation results for chair, table, lamp, and bench.
    }
    \label{fig:qualitative-shapenet}
\end{figure}

\paragraph{Unconditional Results on ShapeNet.} 
We visualize unconditional 3D mesh generation results for chair, table, lamp and bench in \cref{fig:qualitative-shapenet}.
One can see that MeshXL is able to produce diverse and high-quality 3D meshes.

\begin{figure}[htbp]
    \centering
    \makebox[\textwidth][c]{
        \includegraphics[width=1.25\linewidth]{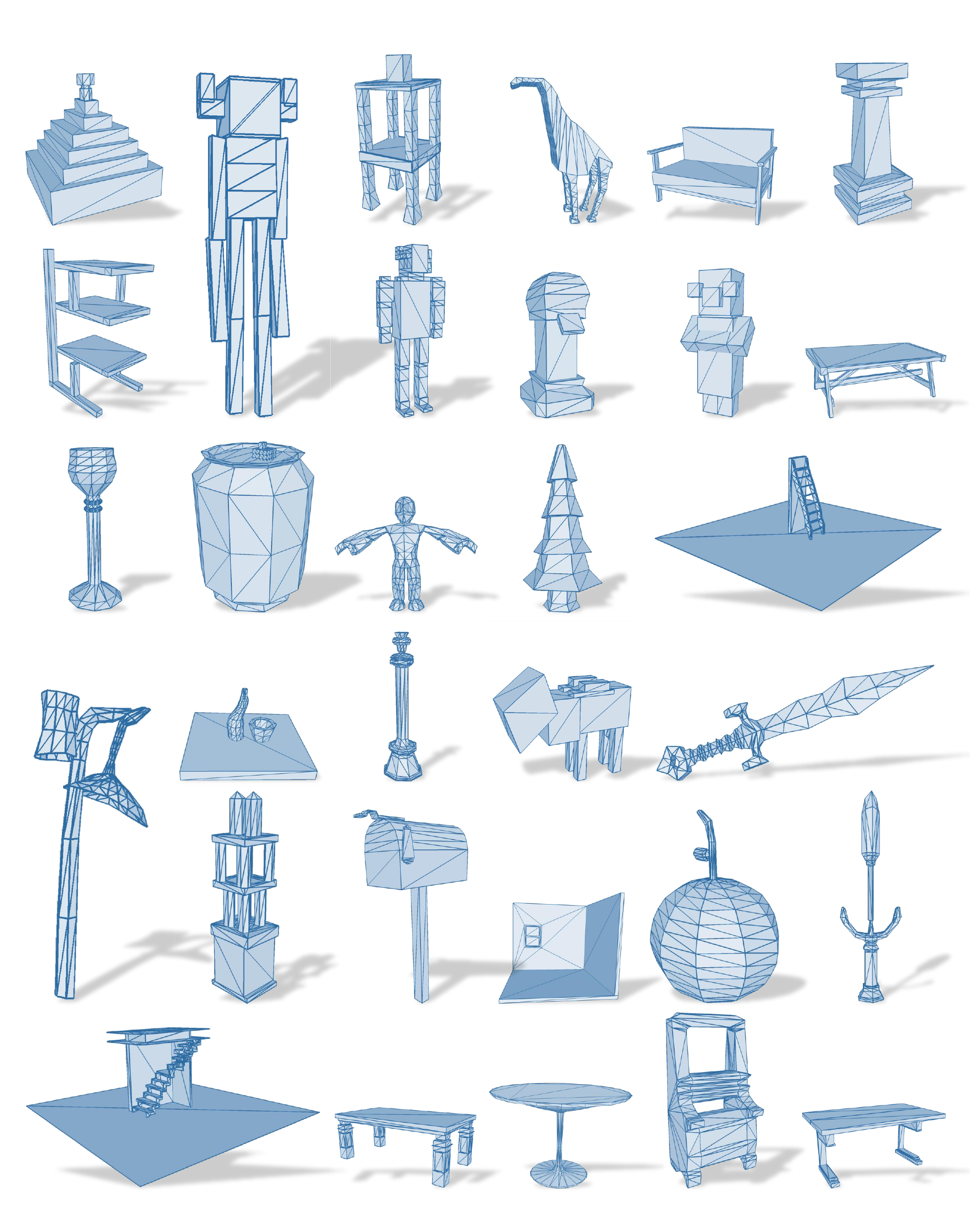}
    }
    \caption{
        \textbf{Gallery results.}
        MeshXL is able to produce diverse 3D meshes with high quality.
    }
    \label{fig:qualitative-random}
\end{figure}

\paragraph{Unconditional Generation on Objaverse.}
We visualize 3D meshes randomly sampled from MeshXL base model in \cref{fig:qualitative-random}.
After training on a large-scale collection of 3D mesh data, MeshXL is able to produce diverse and high-quality 3D meshes.

\section{Discussions}

\paragraph{Difference with PolyGen}~\cite{nash2020polygen}. 
PolyGen explores the auto-regressive generation of 3D polynomial meshes with two transformers~\cite{vaswani2017attention}, \textit{i.e.} the \textit{vertex transformer} and the \textit{face transformer}.
PolyGen first generates a set of points representing the vertices of the 3D meshes with a vertex transformer.
After that, PolyGen inputs the generated point cloud into the face transformer and predicts the connectivity among the generated with a face transformer.
However, our proposed MeshXL is a more straightforward and end-to-end approach that directly generates the polynomial meshes auto-regressively with decoder-only transformers.

\paragraph{Difference with MeshGPT}~\cite{siddiqui2023meshgpt}. 
MeshGPT consists of a mesh VQVAE~\cite{van2017vqvae} and a decoder-only transformer~\cite{radford2019gpt2}.
MeshGPT first learns a mesh VQVAE to quantize the 3D meshes into discrete tokens.
After that, MeshGPT trains a decoder-only transformer to generate the discrete tokens for 3D mesh reconstruction.
In comparison, our proposed MeshXL is an end-to-end method that learns the neural representation of coordinates and outputs 3D meshes directly.

\paragraph{Extensibility.}
Our method, MeshXL, is built upon the concept of auto-regressive methods.
Therefore, our method is not restricted to the decoder-only transformers~\cite{radford2019gpt2,zhang2022opt,touvron2023llama,touvron2023llama2}, and can also be extended to other causal language models (\textit{i.e.} Mamba~\cite{gu2023mamba}, RWKV~\cite{peng2023rwkv}, and xLSTM~\cite{beck2024xlstm}).

\section{Limitations, Future Work, and Conclusions}

\paragraph{Limitations and Future Work.}
The main drawback of MeshXLs is the inference time.
During sampling, MeshXL will generate 7,200 tokens for an 800-faced 3D mesh, which takes a relatively long time because of the auto-regressive process.
As for future works, recent endeavors on the RNN-related methods~\cite{beck2024xlstm,peng2023rwkv,gu2023mamba} and multiple tokens prediction for LLMs~\cite{gloeckle2024better} might open up great opportunities in saving the inference cost.


\paragraph{Conclusion.}
We validate that NeurCF, an explicit coordinate representation with implicit neural embeddings, is a simple-and-effective representation of 3D meshes.
By modelling the 3D mesh generation as an auto-regressive problem, we seek help from modern LLM approaches and present a family of generative pre-trained models, MeshXL, for high-fidelity 3D mesh generation.
We show that MeshXL performs better given larger-scale training data and increased parameters.
Extensive results show our proposed MeshXL can not only generate high-quality 3D meshes, but also exhibits great potential serving as base models for conditional 3D assets generation.



\clearpage
{
    \small
    \bibliographystyle{plain}
    \bibliography{reference}

\begin{thebibliography}{100}

\bibitem{achlioptas2018learning}
Panos Achlioptas, Olga Diamanti, Ioannis Mitliagkas, and Leonidas Guibas.
\newblock Learning representations and generative models for 3d point clouds.
\newblock In {\em International conference on machine learning}, pages 40--49. PMLR, 2018.

\bibitem{alliegro2023polydiff}
Antonio Alliegro, Yawar Siddiqui, Tatiana Tommasi, and Matthias Nie{\ss}ner.
\newblock Polydiff: Generating 3d polygonal meshes with diffusion models.
\newblock {\em arXiv preprint arXiv:2312.11417}, 2023.

\bibitem{armeni20163d}
Iro Armeni, Ozan Sener, Amir~R Zamir, Helen Jiang, Ioannis Brilakis, Martin Fischer, and Silvio Savarese.
\newblock 3d semantic parsing of large-scale indoor spaces.
\newblock In {\em Proceedings of the IEEE conference on computer vision and pattern recognition}, pages 1534--1543, 2016.

\bibitem{austin2021discrete-diffusion}
Jacob Austin, Daniel~D Johnson, Jonathan Ho, Daniel Tarlow, and Rianne Van Den~Berg.
\newblock Structured denoising diffusion models in discrete state-spaces.
\newblock {\em Advances in Neural Information Processing Systems}, 34:17981--17993, 2021.

\bibitem{barron2021mip-nerf}
Jonathan~T Barron, Ben Mildenhall, Matthew Tancik, Peter Hedman, Ricardo Martin-Brualla, and Pratul~P Srinivasan.
\newblock Mip-nerf: A multiscale representation for anti-aliasing neural radiance fields.
\newblock In {\em Proceedings of the IEEE/CVF International Conference on Computer Vision}, pages 5855--5864, 2021.

\bibitem{beck2024xlstm}
Maximilian Beck, Korbinian P{\"o}ppel, Markus Spanring, Andreas Auer, Oleksandra Prudnikova, Michael Kopp, G{\"u}nter Klambauer, Johannes Brandstetter, and Sepp Hochreiter.
\newblock xlstm: Extended long short-term memory.
\newblock {\em arXiv preprint arXiv:2405.04517}, 2024.

\bibitem{behley2019semantickitti}
Jens Behley, Martin Garbade, Andres Milioto, Jan Quenzel, Sven Behnke, Cyrill Stachniss, and Jurgen Gall.
\newblock Semantickitti: A dataset for semantic scene understanding of lidar sequences.
\newblock In {\em Proceedings of the IEEE/CVF international conference on computer vision}, pages 9297--9307, 2019.

\bibitem{cao2024difftf++}
Ziang Cao, Fangzhou Hong, Tong Wu, Liang Pan, and Ziwei Liu.
\newblock Difftf++: 3d-aware diffusion transformer for large-vocabulary 3d generation.
\newblock {\em arXiv preprint arXiv:2405.08055}, 2024.

\bibitem{chang2015shapenet}
Angel~X Chang, Thomas Funkhouser, Leonidas Guibas, Pat Hanrahan, Qixing Huang, Zimo Li, Silvio Savarese, Manolis Savva, Shuran Song, Hao Su, et~al.
\newblock Shapenet: An information-rich 3d model repository.
\newblock {\em arXiv preprint arXiv:1512.03012}, 2015.

\bibitem{chen2023ll3da}
Sijin Chen, Xin Chen, Chi Zhang, Mingsheng Li, Gang Yu, Hao Fei, Hongyuan Zhu, Jiayuan Fan, and Tao Chen.
\newblock Ll3da: Visual interactive instruction tuning for omni-3d understanding, reasoning, and planning.
\newblock {\em arXiv preprint arXiv:2311.18651}, 2023.

\bibitem{chen2023robust}
Zhen Chen, Zherong Pan, Kui Wu, Etienne Vouga, and Xifeng Gao.
\newblock Robust low-poly meshing for general 3d models.
\newblock {\em ACM Transactions on Graphics (TOG)}, 42(4):1--20, 2023.

\bibitem{chen2020bsp}
Zhiqin Chen, Andrea Tagliasacchi, and Hao Zhang.
\newblock Bsp-net: Generating compact meshes via binary space partitioning.
\newblock In {\em Proceedings of the IEEE/CVF conference on computer vision and pattern recognition}, pages 45--54, 2020.

\bibitem{choy2019Minkowski}
Christopher Choy, JunYoung Gwak, and Silvio Savarese.
\newblock 4d spatio-temporal convnets: Minkowski convolutional neural networks.
\newblock In {\em Proceedings of the IEEE Conference on Computer Vision and Pattern Recognition}, pages 3075--3084, 2019.

\bibitem{blender}
Blender~Online Community.
\newblock {\em Blender - a 3D modelling and rendering package}.
\newblock Blender Foundation, Stichting Blender Foundation, Amsterdam, 2018.

\bibitem{dai2017scannet}
Angela Dai, Angel~X Chang, Manolis Savva, Maciej Halber, Thomas Funkhouser, and Matthias Nie{\ss}ner.
\newblock Scannet: Richly-annotated 3d reconstructions of indoor scenes.
\newblock In {\em Proceedings of the IEEE conference on computer vision and pattern recognition}, pages 5828--5839, 2017.

\bibitem{deitke2024objaverse-xl}
Matt Deitke, Ruoshi Liu, Matthew Wallingford, Huong Ngo, Oscar Michel, Aditya Kusupati, Alan Fan, Christian Laforte, Vikram Voleti, Samir~Yitzhak Gadre, et~al.
\newblock Objaverse-xl: A universe of 10m+ 3d objects.
\newblock {\em Advances in Neural Information Processing Systems}, 36, 2024.

\bibitem{deitke2023objaverse}
Matt Deitke, Dustin Schwenk, Jordi Salvador, Luca Weihs, Oscar Michel, Eli VanderBilt, Ludwig Schmidt, Kiana Ehsani, Aniruddha Kembhavi, and Ali Farhadi.
\newblock Objaverse: A universe of annotated 3d objects.
\newblock In {\em Proceedings of the IEEE/CVF Conference on Computer Vision and Pattern Recognition}, pages 13142--13153, 2023.

\bibitem{devlin2018bert}
Jacob Devlin, Ming-Wei Chang, Kenton Lee, and Kristina Toutanova.
\newblock Bert: Pre-training of deep bidirectional transformers for language understanding.
\newblock {\em arXiv preprint arXiv:1810.04805}, 2018.

\bibitem{dosovitskiy2020vit}
Alexey Dosovitskiy, Lucas Beyer, Alexander Kolesnikov, Dirk Weissenborn, Xiaohua Zhai, Thomas Unterthiner, Mostafa Dehghani, Matthias Minderer, Georg Heigold, Sylvain Gelly, et~al.
\newblock An image is worth 16x16 words: Transformers for image recognition at scale.
\newblock {\em arXiv preprint arXiv:2010.11929}, 2020.

\bibitem{driess2023palm}
Danny Driess, Fei Xia, Mehdi~SM Sajjadi, Corey Lynch, Aakanksha Chowdhery, Brian Ichter, Ayzaan Wahid, Jonathan Tompson, Quan Vuong, Tianhe Yu, et~al.
\newblock Palm-e: An embodied multimodal language model.
\newblock {\em arXiv preprint arXiv:2303.03378}, 2023.

\bibitem{fu20213d-front}
Huan Fu, Bowen Cai, Lin Gao, Ling-Xiao Zhang, Jiaming Wang, Cao Li, Qixun Zeng, Chengyue Sun, Rongfei Jia, Binqiang Zhao, et~al.
\newblock 3d-front: 3d furnished rooms with layouts and semantics.
\newblock In {\em Proceedings of the IEEE/CVF International Conference on Computer Vision}, pages 10933--10942, 2021.

\bibitem{fu20213d-future}
Huan Fu, Rongfei Jia, Lin Gao, Mingming Gong, Binqiang Zhao, Steve Maybank, and Dacheng Tao.
\newblock 3d-future: 3d furniture shape with texture.
\newblock {\em International Journal of Computer Vision}, pages 1--25, 2021.

\bibitem{gao2022get3d}
Jun Gao, Tianchang Shen, Zian Wang, Wenzheng Chen, Kangxue Yin, Daiqing Li, Or~Litany, Zan Gojcic, and Sanja Fidler.
\newblock Get3d: A generative model of high quality 3d textured shapes learned from images.
\newblock {\em Advances In Neural Information Processing Systems}, 35:31841--31854, 2022.

\bibitem{gloeckle2024better}
Fabian Gloeckle, Badr~Youbi Idrissi, Baptiste Rozi{\`e}re, David Lopez-Paz, and Gabriel Synnaeve.
\newblock Better \& faster large language models via multi-token prediction.
\newblock {\em arXiv preprint arXiv:2404.19737}, 2024.

\bibitem{goodfellow2020generative}
Ian Goodfellow, Jean Pouget-Abadie, Mehdi Mirza, Bing Xu, David Warde-Farley, Sherjil Ozair, Aaron Courville, and Yoshua Bengio.
\newblock Generative adversarial networks.
\newblock {\em Communications of the ACM}, 63(11):139--144, 2020.

\bibitem{gu2023mamba}
Albert Gu and Tri Dao.
\newblock Mamba: Linear-time sequence modeling with selective state spaces.
\newblock {\em arXiv preprint arXiv:2312.00752}, 2023.

\bibitem{guo2023pointbind_pointllm}
Ziyu Guo, Renrui Zhang, Xiangyang Zhu, Yiwen Tang, Xianzheng Ma, Jiaming Han, Kexin Chen, Peng Gao, Xianzhi Li, Hongsheng Li, et~al.
\newblock Point-bind \& point-llm: Aligning point cloud with multi-modality for 3d understanding, generation, and instruction following.
\newblock {\em arXiv preprint arXiv:2309.00615}, 2023.

\bibitem{ho2020ddpm}
Jonathan Ho, Ajay Jain, and Pieter Abbeel.
\newblock Denoising diffusion probabilistic models.
\newblock {\em Advances in neural information processing systems}, 33:6840--6851, 2020.

\bibitem{hong20243dtopia}
Fangzhou Hong, Jiaxiang Tang, Ziang Cao, Min Shi, Tong Wu, Zhaoxi Chen, Tengfei Wang, Liang Pan, Dahua Lin, and Ziwei Liu.
\newblock 3dtopia: Large text-to-3d generation model with hybrid diffusion priors.
\newblock {\em arXiv preprint arXiv:2403.02234}, 2024.

\bibitem{hong2023lrm}
Yicong Hong, Kai Zhang, Jiuxiang Gu, Sai Bi, Yang Zhou, Difan Liu, Feng Liu, Kalyan Sunkavalli, Trung Bui, and Hao Tan.
\newblock Lrm: Large reconstruction model for single image to 3d.
\newblock {\em arXiv preprint arXiv:2311.04400}, 2023.

\bibitem{hong20233d-llm}
Yining Hong, Haoyu Zhen, Peihao Chen, Shuhong Zheng, Yilun Du, Zhenfang Chen, and Chuang Gan.
\newblock 3d-llm: Injecting the 3d world into large language models.
\newblock {\em Advances in Neural Information Processing Systems}, 36:20482--20494, 2023.

\bibitem{huang2023embodied}
Jiangyong Huang, Silong Yong, Xiaojian Ma, Xiongkun Linghu, Puhao Li, Yan Wang, Qing Li, Song-Chun Zhu, Baoxiong Jia, and Siyuan Huang.
\newblock An embodied generalist agent in 3d world.
\newblock In {\em Proceedings of the International Conference on Machine Learning (ICML)}, 2024.

\bibitem{ibing20213d}
Moritz Ibing, Isaak Lim, and Leif Kobbelt.
\newblock 3d shape generation with grid-based implicit functions.
\newblock In {\em Proceedings of the IEEE/CVF Conference on Computer Vision and Pattern Recognition}, pages 13559--13568, 2021.

\bibitem{jia2024sceneverse}
Baoxiong Jia, Yixin Chen, Huangyue Yu, Yan Wang, Xuesong Niu, Tengyu Liu, Qing Li, and Siyuan Huang.
\newblock Sceneverse: Scaling 3d vision-language learning for grounded scene understanding.
\newblock {\em arXiv preprint arXiv:2401.09340}, 2024.

\bibitem{jiang2023mistral}
Albert~Q Jiang, Alexandre Sablayrolles, Arthur Mensch, Chris Bamford, Devendra~Singh Chaplot, Diego de~las Casas, Florian Bressand, Gianna Lengyel, Guillaume Lample, Lucile Saulnier, et~al.
\newblock Mistral 7b.
\newblock {\em arXiv preprint arXiv:2310.06825}, 2023.

\bibitem{jiang2024motiongpt}
Biao Jiang, Xin Chen, Wen Liu, Jingyi Yu, Gang Yu, and Tao Chen.
\newblock Motiongpt: Human motion as a foreign language.
\newblock {\em Advances in Neural Information Processing Systems}, 36, 2024.

\bibitem{kazhdan2006poisson}
Michael Kazhdan, Matthew Bolitho, and Hugues Hoppe.
\newblock Poisson surface reconstruction.
\newblock In {\em Proceedings of the fourth Eurographics symposium on Geometry processing}, volume~7, 2006.

\bibitem{kerbl20233dgs}
Bernhard Kerbl, Georgios Kopanas, Thomas Leimk{\"u}hler, and George Drettakis.
\newblock 3d gaussian splatting for real-time radiance field rendering.
\newblock {\em ACM Transactions on Graphics}, 42(4):1--14, 2023.

\bibitem{li2023generative}
Chenghao Li, Chaoning Zhang, Atish Waghwase, Lik-Hang Lee, Francois Rameau, Yang Yang, Sung-Ho Bae, and Choong~Seon Hong.
\newblock Generative ai meets 3d: A survey on text-to-3d in aigc era.
\newblock {\em arXiv preprint arXiv:2305.06131}, 2023.

\bibitem{li2023blip2}
Junnan Li, Dongxu Li, Silvio Savarese, and Steven Hoi.
\newblock Blip-2: Bootstrapping language-image pre-training with frozen image encoders and large language models.
\newblock In {\em International conference on machine learning}, pages 19730--19742. PMLR, 2023.

\bibitem{li2023m3dbench}
Mingsheng Li, Xin Chen, Chi Zhang, Sijin Chen, Hongyuan Zhu, Fukun Yin, Gang Yu, and Tao Chen.
\newblock M3dbench: Let's instruct large models with multi-modal 3d prompts.
\newblock {\em arXiv preprint arXiv:2312.10763}, 2023.

\bibitem{li2024advances}
Xiaoyu Li, Qi~Zhang, Di~Kang, Weihao Cheng, Yiming Gao, Jingbo Zhang, Zhihao Liang, Jing Liao, Yan-Pei Cao, and Ying Shan.
\newblock Advances in 3d generation: A survey.
\newblock {\em arXiv preprint arXiv:2401.17807}, 2024.

\bibitem{liu2023one-2-3-45++}
Minghua Liu, Ruoxi Shi, Linghao Chen, Zhuoyang Zhang, Chao Xu, Xinyue Wei, Hansheng Chen, Chong Zeng, Jiayuan Gu, and Hao Su.
\newblock One-2-3-45++: Fast single image to 3d objects with consistent multi-view generation and 3d diffusion.
\newblock {\em arXiv preprint arXiv:2311.07885}, 2023.

\bibitem{liu2024openshape}
Minghua Liu, Ruoxi Shi, Kaiming Kuang, Yinhao Zhu, Xuanlin Li, Shizhong Han, Hong Cai, Fatih Porikli, and Hao Su.
\newblock Openshape: Scaling up 3d shape representation towards open-world understanding.
\newblock {\em Advances in Neural Information Processing Systems}, 36, 2024.

\bibitem{liu2024one-2-3-45}
Minghua Liu, Chao Xu, Haian Jin, Linghao Chen, Mukund Varma~T, Zexiang Xu, and Hao Su.
\newblock One-2-3-45: Any single image to 3d mesh in 45 seconds without per-shape optimization.
\newblock {\em Advances in Neural Information Processing Systems}, 36, 2024.

\bibitem{liu2023zero}
Ruoshi Liu, Rundi Wu, Basile Van~Hoorick, Pavel Tokmakov, Sergey Zakharov, and Carl Vondrick.
\newblock Zero-1-to-3: Zero-shot one image to 3d object.
\newblock In {\em Proceedings of the IEEE/CVF International Conference on Computer Vision}, pages 9298--9309, 2023.

\bibitem{liu2023meshdiffusion}
Zhen Liu, Yao Feng, Michael~J Black, Derek Nowrouzezahrai, Liam Paull, and Weiyang Liu.
\newblock Meshdiffusion: Score-based generative 3d mesh modeling.
\newblock {\em arXiv preprint arXiv:2303.08133}, 2023.

\bibitem{loshchilov2017adamw}
Ilya Loshchilov and Frank Hutter.
\newblock Decoupled weight decay regularization.
\newblock {\em arXiv preprint arXiv:1711.05101}, 2017.

\bibitem{luo2021diffusion}
Shitong Luo and Wei Hu.
\newblock Diffusion probabilistic models for 3d point cloud generation.
\newblock In {\em Proceedings of the IEEE/CVF Conference on Computer Vision and Pattern Recognition}, pages 2837--2845, 2021.

\bibitem{lyu2023controllable}
Zhaoyang Lyu, Jinyi Wang, Yuwei An, Ya~Zhang, Dahua Lin, and Bo~Dai.
\newblock Controllable mesh generation through sparse latent point diffusion models.
\newblock In {\em Proceedings of the IEEE/CVF conference on computer vision and pattern recognition}, pages 271--280, 2023.

\bibitem{mildenhall2020nerf}
Ben Mildenhall, Pratul~P. Srinivasan, Matthew Tancik, Jonathan~T. Barron, Ravi Ramamoorthi, and Ren Ng.
\newblock Nerf: Representing scenes as neural radiance fields for view synthesis.
\newblock In {\em ECCV}, 2020.

\bibitem{mittal2022autosdf}
Paritosh Mittal, Yen-Chi Cheng, Maneesh Singh, and Shubham Tulsiani.
\newblock Autosdf: Shape priors for 3d completion, reconstruction and generation.
\newblock In {\em Proceedings of the IEEE/CVF Conference on Computer Vision and Pattern Recognition}, pages 306--315, 2022.

\bibitem{nash2020polygen}
Charlie Nash, Yaroslav Ganin, SM~Ali Eslami, and Peter Battaglia.
\newblock Polygen: An autoregressive generative model of 3d meshes.
\newblock In {\em International conference on machine learning}, pages 7220--7229. PMLR, 2020.

\bibitem{nichol2022point-e}
Alex Nichol, Heewoo Jun, Prafulla Dhariwal, Pamela Mishkin, and Mark Chen.
\newblock Point-e: A system for generating 3d point clouds from complex prompts.
\newblock {\em arXiv preprint arXiv:2212.08751}, 2022.

\bibitem{peng2023rwkv}
Bo~Peng, Eric Alcaide, Quentin Anthony, Alon Albalak, Samuel Arcadinho, Huanqi Cao, Xin Cheng, Michael Chung, Matteo Grella, Kranthi~Kiran GV, et~al.
\newblock Rwkv: Reinventing rnns for the transformer era.
\newblock {\em arXiv preprint arXiv:2305.13048}, 2023.

\bibitem{poole2022dreamfusion}
Ben Poole, Ajay Jain, Jonathan~T Barron, and Ben Mildenhall.
\newblock Dreamfusion: Text-to-3d using 2d diffusion.
\newblock {\em arXiv preprint arXiv:2209.14988}, 2022.

\bibitem{qi2017pointnet}
Charles~R Qi, Hao Su, Kaichun Mo, and Leonidas~J Guibas.
\newblock Pointnet: Deep learning on point sets for 3d classification and segmentation.
\newblock In {\em Proceedings of the IEEE conference on computer vision and pattern recognition}, pages 652--660, 2017.

\bibitem{qi2017pointnet++}
Charles~Ruizhongtai Qi, Li~Yi, Hao Su, and Leonidas~J Guibas.
\newblock Pointnet++: Deep hierarchical feature learning on point sets in a metric space.
\newblock {\em Advances in neural information processing systems}, 30, 2017.

\bibitem{radford2019gpt2}
Alec Radford, Jeffrey Wu, Rewon Child, David Luan, Dario Amodei, Ilya Sutskever, et~al.
\newblock Language models are unsupervised multitask learners.
\newblock {\em OpenAI blog}, 1(8):9, 2019.

\bibitem{rajbhandari2020zero}
Samyam Rajbhandari, Jeff Rasley, Olatunji Ruwase, and Yuxiong He.
\newblock Zero: Memory optimizations toward training trillion parameter models.
\newblock In {\em SC20: International Conference for High Performance Computing, Networking, Storage and Analysis}, pages 1--16. IEEE, 2020.

\bibitem{ren2023xcube}
Xuanchi Ren, Jiahui Huang, Xiaohui Zeng, Ken Museth, Sanja Fidler, and Francis Williams.
\newblock Xcube ($\mathcal{X}^{3}$): Large-scale 3d generative modeling using sparse voxel hierarchies.
\newblock {\em arXiv preprint arXiv:2312.03806}, 2023.

\bibitem{rombach2022high}
Robin Rombach, Andreas Blattmann, Dominik Lorenz, Patrick Esser, and Bj{\"o}rn Ommer.
\newblock High-resolution image synthesis with latent diffusion models.
\newblock In {\em Proceedings of the IEEE/CVF conference on computer vision and pattern recognition}, pages 10684--10695, 2022.

\bibitem{rombach2022latent-diffusion}
Robin Rombach, Andreas Blattmann, Dominik Lorenz, Patrick Esser, and Bj{\"o}rn Ommer.
\newblock High-resolution image synthesis with latent diffusion models.
\newblock In {\em Proceedings of the IEEE/CVF conference on computer vision and pattern recognition}, pages 10684--10695, 2022.

\bibitem{shen2021dmtet}
Tianchang Shen, Jun Gao, Kangxue Yin, Ming-Yu Liu, and Sanja Fidler.
\newblock Deep marching tetrahedra: a hybrid representation for high-resolution 3d shape synthesis.
\newblock {\em Advances in Neural Information Processing Systems}, 34:6087--6101, 2021.

\bibitem{shue20233d-triplane-diff}
J~Ryan Shue, Eric~Ryan Chan, Ryan Po, Zachary Ankner, Jiajun Wu, and Gordon Wetzstein.
\newblock 3d neural field generation using triplane diffusion.
\newblock In {\em Proceedings of the IEEE/CVF Conference on Computer Vision and Pattern Recognition}, pages 20875--20886, 2023.

\bibitem{siddiqui2023meshgpt}
Yawar Siddiqui, Antonio Alliegro, Alexey Artemov, Tatiana Tommasi, Daniele Sirigatti, Vladislav Rosov, Angela Dai, and Matthias Nie{\ss}ner.
\newblock Meshgpt: Generating triangle meshes with decoder-only transformers.
\newblock {\em arXiv preprint arXiv:2311.15475}, 2023.

\bibitem{song2015sun}
Shuran Song, Samuel~P Lichtenberg, and Jianxiong Xiao.
\newblock Sun rgb-d: A rgb-d scene understanding benchmark suite.
\newblock In {\em Proceedings of the IEEE conference on computer vision and pattern recognition}, pages 567--576, 2015.

\bibitem{tang2024lgm}
Jiaxiang Tang, Zhaoxi Chen, Xiaokang Chen, Tengfei Wang, Gang Zeng, and Ziwei Liu.
\newblock Lgm: Large multi-view gaussian model for high-resolution 3d content creation.
\newblock {\em arXiv preprint arXiv:2402.05054}, 2024.

\bibitem{tang2023dreamgaussian}
Jiaxiang Tang, Jiawei Ren, Hang Zhou, Ziwei Liu, and Gang Zeng.
\newblock Dreamgaussian: Generative gaussian splatting for efficient 3d content creation.
\newblock {\em arXiv preprint arXiv:2309.16653}, 2023.

\bibitem{touvron2023llama}
Hugo Touvron, Thibaut Lavril, Gautier Izacard, Xavier Martinet, Marie-Anne Lachaux, Timoth{\'e}e Lacroix, Baptiste Rozi{\`e}re, Naman Goyal, Eric Hambro, Faisal Azhar, et~al.
\newblock Llama: Open and efficient foundation language models.
\newblock {\em arXiv preprint arXiv:2302.13971}, 2023.

\bibitem{touvron2023llama2}
Hugo Touvron, Louis Martin, Kevin Stone, Peter Albert, Amjad Almahairi, Yasmine Babaei, Nikolay Bashlykov, Soumya Batra, Prajjwal Bhargava, Shruti Bhosale, et~al.
\newblock Llama 2: Open foundation and fine-tuned chat models.
\newblock {\em arXiv preprint arXiv:2307.09288}, 2023.

\bibitem{uy2019scanobjectnn}
Mikaela~Angelina Uy, Quang-Hieu Pham, Binh-Son Hua, Thanh Nguyen, and Sai-Kit Yeung.
\newblock Revisiting point cloud classification: A new benchmark dataset and classification model on real-world data.
\newblock In {\em Proceedings of the IEEE/CVF international conference on computer vision}, pages 1588--1597, 2019.

\bibitem{van2017vqvae}
Aaron Van Den~Oord, Oriol Vinyals, et~al.
\newblock Neural discrete representation learning.
\newblock {\em Advances in neural information processing systems}, 30, 2017.

\bibitem{vaswani2017attention}
Ashish Vaswani, Noam Shazeer, Niki Parmar, Jakob Uszkoreit, Llion Jones, Aidan~N Gomez, {\L}ukasz Kaiser, and Illia Polosukhin.
\newblock Attention is all you need.
\newblock {\em Advances in neural information processing systems}, 30, 2017.

\bibitem{wang2023pf}
Peng Wang, Hao Tan, Sai Bi, Yinghao Xu, Fujun Luan, Kalyan Sunkavalli, Wenping Wang, Zexiang Xu, and Kai Zhang.
\newblock Pf-lrm: Pose-free large reconstruction model for joint pose and shape prediction.
\newblock {\em arXiv preprint arXiv:2311.12024}, 2023.

\bibitem{wang2023embodiedscan}
Tai Wang, Xiaohan Mao, Chenming Zhu, Runsen Xu, Ruiyuan Lyu, Peisen Li, Xiao Chen, Wenwei Zhang, Kai Chen, Tianfan Xue, et~al.
\newblock Embodiedscan: A holistic multi-modal 3d perception suite towards embodied ai.
\newblock {\em arXiv preprint arXiv:2312.16170}, 2023.

\bibitem{wang2023cogvlm}
Weihan Wang, Qingsong Lv, Wenmeng Yu, Wenyi Hong, Ji~Qi, Yan Wang, Junhui Ji, Zhuoyi Yang, Lei Zhao, Xixuan Song, et~al.
\newblock Cogvlm: Visual expert for pretrained language models.
\newblock {\em arXiv preprint arXiv:2311.03079}, 2023.

\bibitem{wang2024prolificdreamer}
Zhengyi Wang, Cheng Lu, Yikai Wang, Fan Bao, Chongxuan Li, Hang Su, and Jun Zhu.
\newblock Prolificdreamer: High-fidelity and diverse text-to-3d generation with variational score distillation.
\newblock {\em Advances in Neural Information Processing Systems}, 36, 2024.

\bibitem{wang2024themestation}
Zhenwei Wang, Tengfei Wang, Gerhard Hancke, Ziwei Liu, and Rynson~W.H. Lau.
\newblock Themestation: Generating theme-aware 3d assets from few exemplars.
\newblock 2024.

\bibitem{wei2024meshlrm}
Xinyue Wei, Kai Zhang, Sai Bi, Hao Tan, Fujun Luan, Valentin Deschaintre, Kalyan Sunkavalli, Hao Su, and Zexiang Xu.
\newblock Meshlrm: Large reconstruction model for high-quality mesh.
\newblock {\em arXiv preprint arXiv:2404.12385}, 2024.

\bibitem{wu2023unleashing}
Hongtao Wu, Ya~Jing, Chilam Cheang, Guangzeng Chen, Jiafeng Xu, Xinghang Li, Minghuan Liu, Hang Li, and Tao Kong.
\newblock Unleashing large-scale video generative pre-training for visual robot manipulation.
\newblock {\em arXiv preprint arXiv:2312.13139}, 2023.

\bibitem{wu2016-3dgan}
Jiajun Wu, Chengkai Zhang, Tianfan Xue, Bill Freeman, and Josh Tenenbaum.
\newblock Learning a probabilistic latent space of object shapes via 3d generative-adversarial modeling.
\newblock {\em Advances in neural information processing systems}, 29, 2016.

\bibitem{wu2015modelnet}
Zhirong Wu, Shuran Song, Aditya Khosla, Fisher Yu, Linguang Zhang, Xiaoou Tang, and Jianxiong Xiao.
\newblock 3d shapenets: A deep representation for volumetric shapes.
\newblock In {\em Proceedings of the IEEE conference on computer vision and pattern recognition}, pages 1912--1920, 2015.

\bibitem{xu2024instantmesh}
Jiale Xu, Weihao Cheng, Yiming Gao, Xintao Wang, Shenghua Gao, and Ying Shan.
\newblock Instantmesh: Efficient 3d mesh generation from a single image with sparse-view large reconstruction models.
\newblock {\em arXiv preprint arXiv:2404.07191}, 2024.

\bibitem{xu2023pointllm}
Runsen Xu, Xiaolong Wang, Tai Wang, Yilun Chen, Jiangmiao Pang, and Dahua Lin.
\newblock Pointllm: Empowering large language models to understand point clouds.
\newblock {\em arXiv preprint arXiv:2308.16911}, 2023.

\bibitem{xu2024grm}
Yinghao Xu, Zifan Shi, Wang Yifan, Hansheng Chen, Ceyuan Yang, Sida Peng, Yujun Shen, and Gordon Wetzstein.
\newblock Grm: Large gaussian reconstruction model for efficient 3d reconstruction and generation.
\newblock {\em arXiv preprint arXiv:2403.14621}, 2024.

\bibitem{xue2023ulip}
Le~Xue, Mingfei Gao, Chen Xing, Roberto Mart{\'\i}n-Mart{\'\i}n, Jiajun Wu, Caiming Xiong, Ran Xu, Juan~Carlos Niebles, and Silvio Savarese.
\newblock Ulip: Learning a unified representation of language, images, and point clouds for 3d understanding.
\newblock In {\em Proceedings of the IEEE/CVF Conference on Computer Vision and Pattern Recognition}, pages 1179--1189, 2023.

\bibitem{xue2023ulip2}
Le~Xue, Ning Yu, Shu Zhang, Junnan Li, Roberto Mart{\'\i}n-Mart{\'\i}n, Jiajun Wu, Caiming Xiong, Ran Xu, Juan~Carlos Niebles, and Silvio Savarese.
\newblock Ulip-2: Towards scalable multimodal pre-training for 3d understanding.
\newblock {\em arXiv preprint arXiv:2305.08275}, 2023.

\bibitem{yeshwanth2023scannet++}
Chandan Yeshwanth, Yueh-Cheng Liu, Matthias Nie{\ss}ner, and Angela Dai.
\newblock Scannet++: A high-fidelity dataset of 3d indoor scenes.
\newblock In {\em Proceedings of the IEEE/CVF International Conference on Computer Vision}, pages 12--22, 2023.

\bibitem{yin2023shapegpt}
Fukun Yin, Xin Chen, Chi Zhang, Biao Jiang, Zibo Zhao, Jiayuan Fan, Gang Yu, Taihao Li, and Tao Chen.
\newblock Shapegpt: 3d shape generation with a unified multi-modal language model.
\newblock {\em arXiv preprint arXiv:2311.17618}, 2023.

\bibitem{yu2023pushing}
Wang Yu, Xuelin Qian, Jingyang Huo, Tiejun Huang, Bo~Zhao, and Yanwei Fu.
\newblock Pushing the limits of 3d shape generation at scale.
\newblock {\em arXiv preprint arXiv:2306.11510}, 2023.

\bibitem{yu2023mvimgnet}
Xianggang Yu, Mutian Xu, Yidan Zhang, Haolin Liu, Chongjie Ye, Yushuang Wu, Zizheng Yan, Chenming Zhu, Zhangyang Xiong, Tianyou Liang, et~al.
\newblock Mvimgnet: A large-scale dataset of multi-view images.
\newblock In {\em Proceedings of the IEEE/CVF conference on computer vision and pattern recognition}, pages 9150--9161, 2023.

\bibitem{zeng2023paint3d}
Xianfang Zeng.
\newblock Paint3d: Paint anything 3d with lighting-less texture diffusion models.
\newblock {\em arXiv preprint arXiv:2312.13913}, 2023.

\bibitem{zhang2022pointclip}
Renrui Zhang, Ziyu Guo, Wei Zhang, Kunchang Li, Xupeng Miao, Bin Cui, Yu~Qiao, Peng Gao, and Hongsheng Li.
\newblock Pointclip: Point cloud understanding by clip.
\newblock In {\em Proceedings of the IEEE/CVF conference on computer vision and pattern recognition}, pages 8552--8562, 2022.

\bibitem{zhang2022opt}
Susan Zhang, Stephen Roller, Naman Goyal, Mikel Artetxe, Moya Chen, Shuohui Chen, Christopher Dewan, Mona Diab, Xian Li, Xi~Victoria Lin, et~al.
\newblock Opt: Open pre-trained transformer language models.
\newblock {\em arXiv preprint arXiv:2205.01068}, 2022.

\bibitem{zhao2024michelangelo}
Zibo Zhao, Wen Liu, Xin Chen, Xianfang Zeng, Rui Wang, Pei Cheng, Bin Fu, Tao Chen, Gang Yu, and Shenghua Gao.
\newblock Michelangelo: Conditional 3d shape generation based on shape-image-text aligned latent representation.
\newblock {\em Advances in Neural Information Processing Systems}, 36, 2024.

\bibitem{zhen20243d-vla}
Haoyu Zhen, Xiaowen Qiu, Peihao Chen, Jincheng Yang, Xin Yan, Yilun Du, Yining Hong, and Chuang Gan.
\newblock 3d-vla: A 3d vision-language-action generative world model.
\newblock {\em arXiv preprint arXiv:2403.09631}, 2024.

\bibitem{zhou2023uni3d}
Junsheng Zhou, Jinsheng Wang, Baorui Ma, Yu-Shen Liu, Tiejun Huang, and Xinlong Wang.
\newblock Uni3d: Exploring unified 3d representation at scale.
\newblock {\em arXiv preprint arXiv:2310.06773}, 2023.

\bibitem{zhou2021pvd}
Linqi Zhou, Yilun Du, and Jiajun Wu.
\newblock 3d shape generation and completion through point-voxel diffusion.
\newblock In {\em Proceedings of the IEEE/CVF International Conference on Computer Vision (ICCV)}, pages 5826--5835, October 2021.

\bibitem{zhu2023ponderv2}
Haoyi Zhu, Honghui Yang, Xiaoyang Wu, Di~Huang, Sha Zhang, Xianglong He, Tong He, Hengshuang Zhao, Chunhua Shen, Yu~Qiao, et~al.
\newblock Ponderv2: Pave the way for 3d foundataion model with a universal pre-training paradigm.
\newblock {\em arXiv preprint arXiv:2310.08586}, 2023.

\bibitem{zhu2023hifa}
Joseph Zhu and Peiye Zhuang.
\newblock Hifa: High-fidelity text-to-3d with advanced diffusion guidance.
\newblock {\em arXiv preprint arXiv:2305.18766}, 2023.

\bibitem{zhu2023pointclipv2}
Xiangyang Zhu, Renrui Zhang, Bowei He, Ziyu Guo, Ziyao Zeng, Zipeng Qin, Shanghang Zhang, and Peng Gao.
\newblock Pointclip v2: Prompting clip and gpt for powerful 3d open-world learning.
\newblock In {\em Proceedings of the IEEE/CVF International Conference on Computer Vision}, pages 2639--2650, 2023.

\end{thebibliography}
}

\end{document}